\pdfoutput=1
\documentclass{article}

\usepackage{microtype}
\usepackage{graphicx}
\usepackage{booktabs}
\usepackage{multirow}
\usepackage{enumitem}
\usepackage{xspace}
\usepackage{float}
\usepackage{placeins}
\usepackage{hyperref}

\usepackage[preprint]{icml2027}
\makeatletter\renewcommand{\ICML@preprint}{\textit{Preprint.}}\makeatother
\hypersetup{pdfauthor={Weiping Yan},pdftitle={STRATA: Symmetry-Aware Physics-Manifold Discovery and Submodular Pareto Labeling on Unlabeled Circuit Graphs}}

\usepackage{amsmath,amssymb,mathtools,amsthm,bm}

\theoremstyle{plain}
\newtheorem{proposition}{Proposition}

\newcommand{\strata}{\textsc{STRATA}\xspace}
\newcommand{\ctpr}{\textsc{CTPR}\xspace}
\newcommand{\pmci}{\textsc{PMCI}\xspace}
\newcommand{\phta}{\textsc{PHTA}\xspace}
\newcommand{\R}{\mathbb{R}}

\newcommand{\TotalRawCircuits}{4,084}
\newcommand{\AMSNetRawCount}{734}
\newcommand{\AnalogGenieRawCount}{3,350}
\newcommand{\UniqueWLTopologies}{3,850}
\newcommand{\DuplicatesRemoved}{234}
\newcommand{\CrossCorpusOverlaps}{27}
\newcommand{\VerifiedAtlasTopologies}{824}
\newcommand{\TotalAtlasArchetypes}{2,850}
\newcommand{\MultiPDKTotalCards}{1,103}

\newcommand{\MeanOrbitCompression}{1.89}
\newcommand{\SkyCohortVerified}{134}
\newcommand{\PtmFortyFiveCohortVerified}{145}
\newcommand{\TotalSpiceEvals}{665,017}
\newcommand{\TotalNgspiceProcs}{80,586}
\newcommand{\HumanN}{330}
\newcommand{\StrataHumanFOne}{0.302}
\newcommand{\StrataHumanFOneFour}{0.3022}
\newcommand{\StrataHumanAcc}{0.273}

\newcommand{\StrataPortlessFOne}{0.0636}
\newcommand{\AmsqReproHumanFOne}{0.054}
\newcommand{\AmsqEnhHumanFOne}{0.065}
\newcommand{\RuleHumanFOne}{0.468}
\newcommand{\WlRgcnHumanFOne}{0.597}
\newcommand{\HumanGainOverEnhX}{4.66}
\newcommand{\HumanGainOverReproX}{5.57}
\newcommand{\StrataHumanEvals}{16,078}
\newcommand{\AmsqEnhHumanEvals}{34,758}
\newcommand{\CrossAmsSkyFOne}{0.3244}
\newcommand{\CrossAmsSkyFOneThree}{0.324}
\newcommand{\CrossAmsPtmFOne}{0.2941}
\newcommand{\CrossAmsPtmFOneThree}{0.294}

\newcommand{\CrossAmsGainX}{6.25}
\newcommand{\CrossAmsSamePdkGainX}{5.67}
\newcommand{\AmsFullStrataSkyFOne}{0.4750}

\newcommand{\AmsFourWayStrataSkyFOne}{0.6068}
\newcommand{\AmsFourWayQReproFOne}{0.2829}

\newcommand{\AmsVerStrataSky}{153}

\newcommand{\StratStrataSkyFOne}{0.4794}

\newcommand{\StratQReproSkyFOne}{0.0994}

\newcommand{\StratSameSkyGainX}{4.82}
\newcommand{\StratSamePtmGainX}{6.14}
\newcommand{\BudgetScaleMinPct}{32.5}
\newcommand{\BudgetScaleMaxPct}{47.9}
\newcommand{\PortRoleAccSky}{69.22}
\newcommand{\PolarityConsistPct}{94.59}
\newcommand{\MeanPrunedClasses}{8.91}
\newcommand{\VisRawBfsFOne}{0.4467}
\newcommand{\VisRawSinaFOne}{0.4451}

\newcommand{\VisRawStrataCtprFOne}{0.4902}

\newcommand{\HealRawCompilePct}{58.95}
\newcommand{\HealCtprCompilePct}{100.0}
\newcommand{\HealRawFuncPct}{23.16}
\newcommand{\HealCtprFuncPct}{76.84}
\newcommand{\HealDsOrientPct}{95.31}

\newcommand{\PhtaFFR}{0.00}
\newcommand{\AmsqFFR}{13.75}
\newcommand{\PhtaHDVR}{3.12}
\newcommand{\AmsqHDVR}{16.25}
\newcommand{\PhtaNHV}{0.2075}

\newcommand{\ParetoKmNHV}{0.1646}
\newcommand{\PhtaPOne}{40.0}
\newcommand{\AmsqPOne}{18.3}
\newcommand{\ParetoKmPOne}{28.6}

\newcommand{\PhtaPvtFifteen}{84.07}
\newcommand{\PhtaPvtFortyFive}{83.09}
\newcommand{\PhtaPelgromYield}{83.33}
\newcommand{\PhtaVosMv}{26.06}
\newcommand{\SamePoolAmsqPvtFifteen}{80.00}
\newcommand{\SamePoolAmsqPvtFortyFive}{78.77}
\newcommand{\SamePoolAmsqVosMv}{17.17}
\newcommand{\EteAmsqPvtFifteen}{5.56}

\newcommand{\AblBaseFOne}{0.1420}
\newcommand{\AblErcFOne}{0.1703}
\newcommand{\AblProbeFOne}{0.1674}
\newcommand{\AblTierOneFOne}{0.4458}
\newcommand{\AblOrbitsFOne}{0.5486}
\newcommand{\AblFullFOne}{0.5774}
\newcommand{\AblLooNoProbeFOne}{0.5995}
\newcommand{\AblLooNoTierOneFOne}{0.1464}

\newcommand{\LamZeroStrictHdvr}{54.0}
\newcommand{\LamOneStrictHdvr}{40.0}
\newcommand{\LamStdStrictHdvr}{10.5}
\newcommand{\LamSixStrictHdvr}{2.5}
\newcommand{\LamStdUnfiltNHV}{0.0556}
\newcommand{\KFiveHvRetain}{86.54}
\newcommand{\XferTwentyeightToFortyfiveZS}{53.3}
\newcommand{\XferTwentyeightToFortyfiveWarm}{65.0}

\newcommand{\WarmColdEvals}{45.16}

\newcommand{\WarmRandEvals}{41.04}
\newcommand{\WarmWlEvals}{38.52}
\newcommand{\WarmAmsqEvals}{37.32}

\newcommand{\WarmStrataEvals}{34.60}
\newcommand{\WarmStrataClosure}{66.0}
\newcommand{\WarmStrataSpeedup}{1.31}
\newcommand{\WarmWilcoxonP}{0.0126}
\newcommand{\WarmRandP}{0.176}
\newcommand{\WarmWlP}{0.238}
\newcommand{\WarmAmsqP}{0.238}
\newcommand{\EnvSingleRTwo}{-0.7936}
\newcommand{\EnvAmsqUnfiltRTwo}{0.3587}
\newcommand{\EnvAmsqFeasRTwo}{0.4416}
\newcommand{\EnvStrataRTwo}{0.4041}
\newcommand{\EnvFullAtlasRTwo}{0.5054}
\newcommand{\CparZeroGbw}{91.47}
\newcommand{\CparHundredGbw}{71.10}
\newcommand{\CparHundredRetain}{95.83}

\icmltitlerunning{STRATA: Symmetry-Aware Physics-Manifold Discovery and Submodular Pareto Labeling on Unlabeled Circuit Graphs}

\begin{document}

\twocolumn[
  \icmltitle{STRATA: Symmetry-Aware Physics-Manifold Discovery and \\Submodular Pareto Labeling on Unlabeled Circuit Graphs}

  \begin{icmlauthorlist}
    \icmlauthor{Weiping Yan}{umncse,umnece}
  \end{icmlauthorlist}
  \icmlaffiliation{umncse}{Department of Computer Science and Engineering, University of Minnesota, Minneapolis, MN, USA.}
  \icmlaffiliation{umnece}{Department of Electrical and Computer Engineering, University of Minnesota, Minneapolis, MN, USA}
  \icmlkeywords{Graph Learning, Scientific Machine Learning, Electronic Design Automation, Submodular Optimization, Pareto Frontier, Analog Circuits}
  \vskip 0.28in
]
\makeatletter\gdef\icmlcorrespondingauthor@text{Weiping Yan (University of Minnesota)}\makeatother
\printAffiliationsAndNotice{}

\begin{abstract}
Learning generative and predictive models on circuit graphs requires datasets pairing topologies with verified trade-off manifolds, yet public analog corpora provide only unlabeled, defect-prone netlists. Existing unsupervised pipelines suffer verification-blind corpus collapse: syntax and polarity ambiguities break differential equilibrium, symmetry-blind per-device sizing fails on multi-stage graphs, and $\max(\mathbf{Z},\mathbf{0})$ clipping masks deficits inside $K$-means centroids. We present \strata, an unsupervised physics-verified framework that canonicalizes circuit graphs, identifies classes across ten analog families, and distills each feasible graph into a Pareto trade-off card with \texttt{SKY130} multi-corner amplifier certification. Canonical Topology and Port Recovery (\ctpr) heals orientation defects via Laplacian potentials, extracts 1-WL symmetry orbits, and checks polarity and servoed common-mode observability in one solve, enabling Physics-Manifold Circuit Identification (\pmci) to pair zero-SPICE pruning with orbit-level $g_m/I_D$ lookup sizing. Pareto-Hypervolume Trade-Off Atlas (\phta) combines strict Pareto filtering, a bi-Lipschitz homeomorphism for deficit fallback, and $(1-1/e)$-optimal submodular hypervolume selection. Across \TotalRawCircuits{} graphs and \TotalSpiceEvals{} \texttt{ngspice-42} evaluations on 130nm/45nm/28nm CMOS, \strata lifts SPICE-verified 11-class Macro-F1 by \HumanGainOverEnhX{}$\times$--\StratSamePtmGainX{}$\times$ on matched 28nm (\StratSameSkyGainX{}$\times$ on \texttt{SKY130}), cuts false-frontier medoids to \PhtaFFR{}\% (vs.\ \AmsqFFR{}\%) while lifting $k=1$ Precision@1 to \PhtaPOne{}\% (vs.\ \AmsqPOne{}\%), and yields \VerifiedAtlasTopologies{} verified topologies (\MultiPDKTotalCards{} multi-PDK cards) speeding OpAmp sizing by \WarmStrataSpeedup{}$\times$ over cold-start ($p_{\mathrm{Holm}}=\WarmWilcoxonP{}$).
\end{abstract}

\section{INTRODUCTION}
\label{sec:intro}

\begin{figure}[t]
  \centering
  \includegraphics[width=\columnwidth]{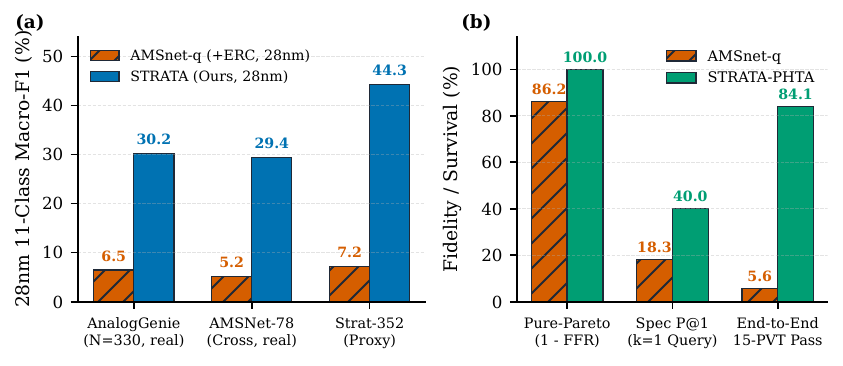}
  \caption{\textbf{Overcoming Verification-Blind Corpus Collapse on Circuit Graphs.} (a)~On strictly matched 28nm CMOS given blind graphs with unlabeled boundary pins, \strata lifts SPICE-verified 11-class Macro-F1 by \HumanGainOverEnhX{}$\times$ on pure-human \texttt{AnalogGenie} ($N=\HumanN{}$), \CrossAmsSamePdkGainX{}$\times$ on cross-corpus \texttt{AMSNet-78}, and \StratSamePtmGainX{}$\times$ on \texttt{Stratified-352} (\StratSameSkyGainX{}$\times$ on matched \texttt{SKY130}). (b)~On dense \texttt{SKY130} pools, \phta eliminates false-frontier medoids (\PhtaFFR{}\% FFR), more than doubles $k=1$ query Precision@1 (\PhtaPOne{}\% vs.\ \AmsqPOne{}\%), and lifts end-to-end 15-corner PVT survival ($N=18$ amplifiers) from \EteAmsqPvtFifteen{}\% to \PhtaPvtFifteen{}\%.}
  \label{fig:teaser}
\end{figure}

Scientific machine learning on structured engineering systems---spanning analog circuit synthesis~\citep{chang2024lamagic,gao2025analoggenie,lai2025analogcoder} and transferable graph-based optimization~\citep{settaluri2020autockt,wang2020gcnrl,dong2023cktgnn}---faces a fundamental supervision bottleneck. Unlike molecular or vision benchmarks where static labels suffice, the semantics of an analog circuit graph are governed by nonlinear differential-algebraic equations (BSIM4 SPICE physics): whether a bipartite device-net graph functions as a stable operational amplifier, comparator, voltage regulator, or oscillator depends on the existence of a symmetry-consistent parameter vector $\bm{\theta}$ satisfying coupled multi-objective constraints~\citep{razavi2016design,jespers2017systematic}.

Recent dataset efforts have mined thousands of raw schematic diagrams and SPICE netlists, including \texttt{AMSNet~1.0}~\citep{tao2024amsnet}, \texttt{AMSnet~2.0}~\citep{shi2025amsnet2}, \texttt{AMSnet-KG}~\citep{shi2024amsnetkg}, \texttt{Masala-CHAI}~\citep{bhandari2024masalachai}, \texttt{SINA}~\citep{aldowaish2026sina}, \texttt{AnalogGenie}~\citep{gao2025analoggenie}, and \texttt{AnalogRetriever}~\citep{wang2026analogretriever}. Because raw netlists omit port polarities, device dimensions, and achievable trade-offs, \citet{zhang2026amsnetq} proposed \texttt{AMSnet-q} to identify and label circuits unsupervised via black-box SPICE sizing and clipped $K$-means clustering across four coarse classes.

When evaluated across multi-corpus repositories and ten functional circuit families, existing unsupervised pipelines suffer from \emph{verification-blind corpus collapse} (Figure~\ref{fig:teaser}) due to three coupled failure modes:
(1)~\emph{Graph dialect and feedback-polarity ambiguity}: 5-node MOSFET syntax errors, floating substrate wells, and inverted drain/source terminals on symmetric schematic glyphs prevent simulator compilation, while unlabeled differential ports leave inverting/non-inverting signs and common-mode feedback (CMFB) demands unresolved;
(2)~\emph{Unpruned, symmetry-breaking search explosion}: testing incompatible classes and optimizing independent $(W_i, L_i)$ variables per transistor dilutes simulation budgets and violates automorphic symmetry across matched differential pairs and current mirrors, stranding multi-stage circuits in cut-off or triode regimes (\AmsqEnhHumanFOne{} Macro-F1 even at $2.25\times$ simulation budget); and
(3)~\emph{One-sided $\max(\mathbf{Z},\mathbf{0})$ deficit collapse}: truncating negative normalized margins onto zero destroys strict Pareto dominance, yielding \AmsqFFR{}\% Pareto-dominated medoids and \AmsqHDVR{}\% hidden specification violations.

To resolve these challenges, we propose \strata (\textbf{S}imulation-Verified \textbf{T}opology \textbf{R}ecovery \textbf{a}nd \textbf{T}rade-Off \textbf{A}tlas). Starting from blind bipartite circuit graphs with stripped identifiers and unlabeled boundary terminals, \strata integrates three modules. \emph{Canonical Topology and Port Recovery} (\ctpr) repairs electrical-rule and drain/source orientation errors via a grounded Laplacian potential solver, extracts 1-WL automorphic symmetry orbits (\MeanOrbitCompression{}$\times$ dimensionality reduction), and checks low-frequency differential polarity and servoed CMFB observability in one linearized solve (Proposition~\ref{prop:single_probe}). \emph{Physics-Manifold Circuit Identification} (\pmci) prunes \MeanPrunedClasses{} of 10 structurally incompatible classes at zero simulation cost and optimizes orbit-level $(g_m/I_D, \log_{10}I_D, L)$ coordinates through foundry BSIM4 lookup manifolds~\citep{jespers2017systematic}. \emph{Pareto-Hypervolume Trade-Off Atlas} (\phta) prioritizes strictly feasible Pareto frontiers, applies a bi-Lipschitz deficit-penalized homeomorphism $\psi_\lambda$ for partial-feasibility fallback, and selects $K$ diverse archetypes via $(1-1/e)$-optimal greedy submodular hypervolume maximization (Proposition~\ref{prop:phta}), with multi-corner PVT and Pelgrom mismatch verification~\citep{pelgrom1989matching} on foundry \texttt{SKY130} amplifiers.

Our contributions are summarized as follows:
\begin{itemize}[leftmargin=1.35em,itemsep=1.5pt,topsep=2pt]
  \item \textbf{Symmetry-Aware Recovery and Hierarchical Physics Identification:} We introduce \ctpr and \pmci, establishing linearized single-probe polarity (\PolarityConsistPct{}\% sign consistency across active differential pairs) and servoed CMFB observability (Proposition~\ref{prop:single_probe}), and combining Tier-1 pruning with symmetry-orbit BSIM4 $g_m/I_D$ sizing to boost blind simulation-verified 11-class Macro-F1 by \HumanGainOverEnhX{}$\times$--\StratSamePtmGainX{}$\times$ on matched 28nm CMOS (\StratSameSkyGainX{}$\times$ on matched \texttt{SKY130}).
  \item \textbf{Bi-Lipschitz Submodular Pareto Labeling:} We develop \phta with provable bi-Lipschitz Pareto preservation for deficit fallback and $(1-1/e)$ submodular hypervolume guarantees (Proposition~\ref{prop:phta}), cutting false-frontier contamination from \AmsqFFR{}\% to \PhtaFFR{}\% and lifting $k=1$ query Precision@1 from \AmsqPOne{}\% to \PhtaPOne{}\% ($+11.4$ points over Pareto-only Euclidean $K$-means).
  \item \textbf{Multi-PDK \strata-Atlas and Topology-Disjoint Transfer:} Auditing \TotalRawCircuits{} public graphs via \TotalSpiceEvals{} \texttt{ngspice-42} BSIM4 evaluations across 130nm, 45nm, and 28nm CMOS, we release \VerifiedAtlasTopologies{} verified topologies (\MultiPDKTotalCards{} multi-PDK cards) that accelerate 5-fold topology-disjoint OpAmp sizing by \WarmStrataSpeedup{}$\times$ over cold-start CMA-ES ($p_{\mathrm{Holm}}=\WarmWilcoxonP{}$) and lift end-to-end blind-graph envelope regression $R^2$ from \EnvSingleRTwo{} ($k=1$) and \EnvAmsqUnfiltRTwo{} (unfiltered \texttt{AMSnet-q}) to \EnvStrataRTwo{}.
\end{itemize}

\section{RELATED WORK}
\label{sec:related}

\noindent\textbf{Circuit Graph Corpora and Schematic Parsing.}
Graph benchmarks for analog circuits initially targeted fixed-topology sizing~\citep{settaluri2020autockt,wang2020gcnrl} or synthetic OpAmp graph generation~\citep{dong2023cktgnn}. Recent schematic-to-netlist pipelines---\texttt{AMSNet~1.0}~\citep{tao2024amsnet}, \texttt{AMSnet~2.0}~\citep{shi2025amsnet2}, \texttt{AMSnet-KG}~\citep{shi2024amsnetkg}, \texttt{Masala-CHAI}~\citep{bhandari2024masalachai}, and \texttt{SINA}~\citep{aldowaish2026sina}---combine symbol detection with pixel, segmentation, or LLM-assisted wire linking, while \texttt{AnalogGenie}~\citep{gao2025analoggenie} assembles human-categorized subcircuit netlists. However, raw corpora lack device sizings and contain syntax defects, inverted symmetric drain/source terminals, and \DuplicatesRemoved{} cross-corpus isomorphic duplicates, motivating \ctpr's canonical 1-WL deduplication and repair.

\smallskip
\noindent\textbf{Unsupervised Graph Identification and Multi-Objective Labeling.}
Relational GNNs~\citep{schlichtkrull2018modeling} and cross-modal retrievers~\citep{wang2026analogretriever} encode circuit graphs without verifying physical specification closure. \citet{zhang2026amsnetq} (\texttt{AMSnet-q}) introduced SPICE-in-the-loop identification and $\max(\mathbf{Z},\mathbf{0})$ $K$-means labeling across four classes, yet independent per-device search breaks automorphic symmetry and zero-clipping masks sub-threshold deficits. Drawing on $g_m/I_D$ inversion-manifold design~\citep{jespers2017systematic} across CMOS PDKs~\citep{skywater2020pdk,zhao2006new}, evolutionary optimization~\citep{hansen2001completely,deb2002fast}, hypervolume indicators~\citep{zitzler1999multiobjective}, and monotone submodular maximization~\citep{nemhauser1978analysis}, \strata scales physical graph discovery to 10 AMS families with provable Pareto guarantees (Table~\ref{tab:competitors}).

\begin{table*}[t]
  \centering
  \caption{\textbf{Comparison of \strata Against Recent AMS Circuit Corpora and Unsupervised Pipelines.} $^\S$\texttt{AnalogGenie} provides unsized topologies with chapter-level categories; \strata's \MultiPDKTotalCards{} total cards span 28nm (\VerifiedAtlasTopologies{}), 45nm (\PtmFortyFiveCohortVerified{}), and 130nm (\SkyCohortVerified{}), with PVT/MC certified on $N=18$ \texttt{SKY130} amplifiers.}
  \label{tab:competitors}
  \resizebox{0.98\textwidth}{!}{\begin{tabular}{lccccccc}
\toprule
\textbf{Corpus / Framework} & \textbf{Venue / Year} & \textbf{Raw / Ver.\ Topos} & \textbf{Classes} & \textbf{CMOS PDKs} & \textbf{ERC + Symmetry} & \textbf{Labeling Geometry} & \textbf{PVT \& Pelgrom} \\
\midrule
AMSNet 1.0~\cite{tao2024amsnet} & arXiv '24 & \AMSNetRawCount{} / 0 & Unlabeled & None & $\times$ & Unsized connectivity & $\times$ \\
Masala-CHAI~\cite{bhandari2024masalachai} & arXiv '24 & LLM-mined / Partial & Mixed LLM & Single & LLM syntax only & Nominal text / SPICE & $\times$ \\
AnalogGenie~\cite{gao2025analoggenie} & ICLR '25 & \AnalogGenieRawCount{} / 0$^\S$ & Chapter tags & Single & $\times$ & Unsized pin-graph & $\times$ \\
AMSnet 2.0~\cite{shi2025amsnet2} & arXiv '25 & Schematics / 0 & Unlabeled & None & AI net seg.\ only & Unsized connectivity & $\times$ \\
SINA~\cite{aldowaish2026sina} & arXiv '26 & Schematics / 0 & Unlabeled & None & CCL wire link & Unsized netlist & $\times$ \\
AMSnet-q~\cite{zhang2026amsnetq} & arXiv '26 & 739 / 105 & 4 coarse & 1 (28nm) & $\times$ (Per-MOS) & $\max(\mathbf{Z},\mathbf{0})$ $K$-Means & $\times$ \\
\midrule
\textbf{\strata (Ours)} & \textbf{2027} & \textbf{\TotalRawCircuits{} / \VerifiedAtlasTopologies{} (\MultiPDKTotalCards{})} & \textbf{10 AMS} & \textbf{3 (130/45/28nm)} & \textbf{\ctpr + 1-WL Orbits} & \textbf{Bi-Lipschitz \phta ($1-1/e$)} & \textbf{SKY130 15/45-PVT + MC} \\
\bottomrule
\end{tabular}
}
\end{table*}

\section{METHOD}
\label{sec:method}

\begin{figure*}[t]
  \centering
  \includegraphics[width=0.98\textwidth]{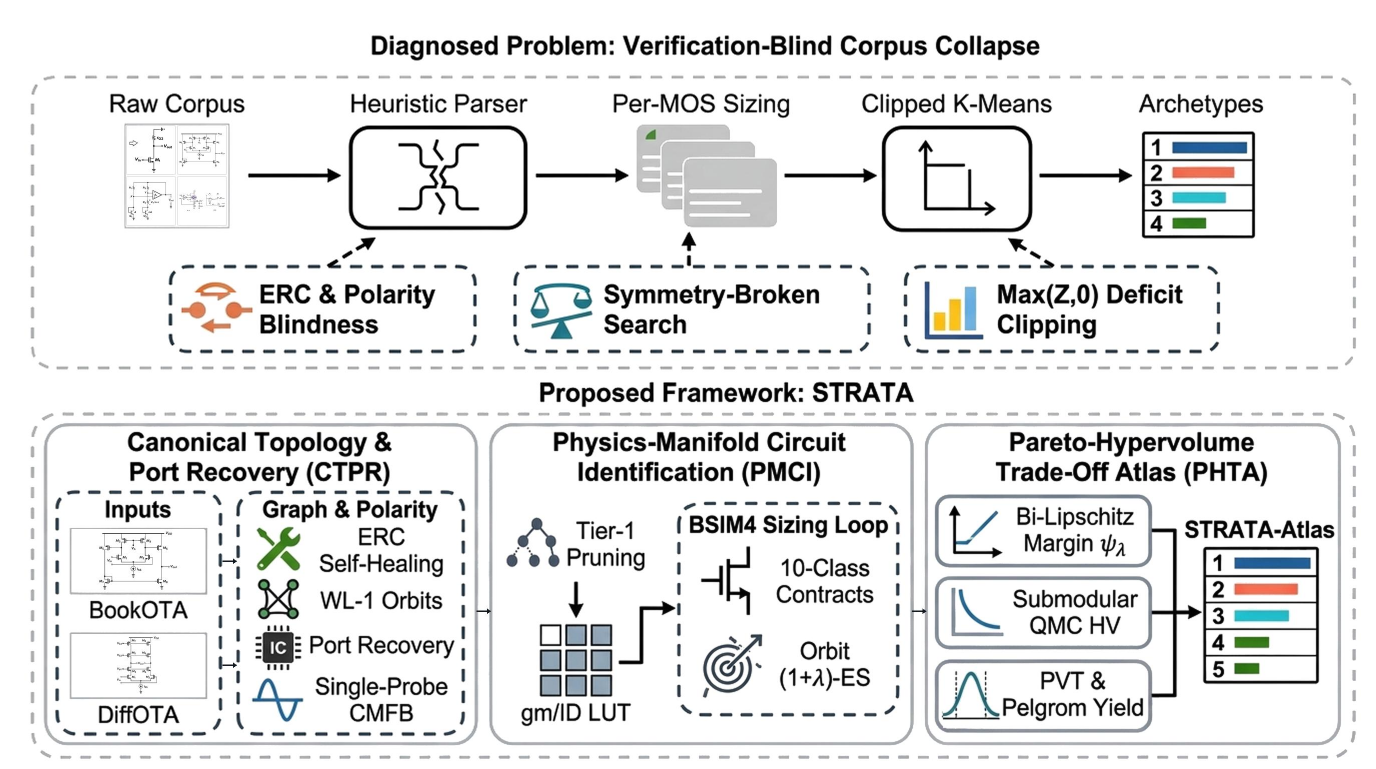}
  \caption{\textbf{Overview of \strata.} Contrasted against verification-blind pipelines (top), \strata processes real schematic graphs and blind netlists via three coupled modules (bottom): (1)~\textbf{\ctpr} repairs ERC/orientation defects, extracts 1-WL symmetry orbits, and checks port polarity and servoed CMFB observability in one linearized probe; (2)~\textbf{\pmci} prunes structurally incompatible classes at zero SPICE cost and searches orbit-level $(g_m/I_D, \log_{10} I_D, L)$ manifolds through BSIM4 lookup tables across 10 AMS classes; and (3)~\textbf{\phta} combines strict Pareto filtering with bi-Lipschitz signed margins $\psi_\lambda$, submodular QMC hypervolume selection, and \texttt{SKY130} PVT/Pelgrom certification.}
  \label{fig:framework}
\end{figure*}

\subsection{Problem Formulation and Overview}
Let $\mathcal{D} = \{G_n\}_{n=1}^N$ be an unlabeled collection of raw analog netlists or schematic wire graphs. To guarantee zero metadata leakage, every circuit is sanitized into a blind pin-typed bipartite graph $G = (\mathcal{V}_{\mathrm{dev}} \cup \mathcal{V}_{\mathrm{net}}, \mathcal{E}, \tau, \mathcal{P}_{\mathrm{ext}})$, where $\mathcal{V}_{\mathrm{dev}}$ contains shuffled device nodes $\{D_0, \dots, D_{|\mathcal{V}_{\mathrm{dev}}|-1}\}$ typed by $\tau(D_i) \in \{\text{nmos}, \text{pmos}, R, C, L, D, Q\}$, $\mathcal{V}_{\mathrm{net}}$ contains supply rails $\{\mathrm{VDD}, \mathrm{VSS}\}$ plus shuffled opaque nets $\{n_0, \dots, n_k\}$, $\mathcal{E}$ encodes terminal-role edges $(D_i, p, n_r)$ for $p \in \{\mathrm{D}, \mathrm{G}, \mathrm{S}, \mathrm{B}, \mathrm{P}, \mathrm{N}\}$, and $\mathcal{P}_{\mathrm{ext}} \subseteq \mathcal{V}_{\mathrm{net}} \setminus \{\mathrm{VDD}, \mathrm{VSS}\}$ lists unlabeled boundary terminal pins.

Given $G$ and foundry BSIM4 process decks $\mathcal{T}$ (130nm, 45nm, 28nm), \strata outputs: (i)~a canonical repaired graph $\widehat{G}$ with port-role assignment $\widehat{\rho}$; (ii)~a verified functional class $\widehat{c} \in \mathcal{C} \cup \{\text{Unclassified}\}$ across $|\mathcal{C}|=10$ AMS families (abstaining as \text{Unclassified} when no physically feasible parameter vector exists); and (iii)~a trade-off card $\mathcal{A}(\widehat{G}, \widehat{c})$ comprising $K_{\mathrm{eff}} = \min(K, |\mathcal{Y}|)$ ($K \in \mathbb{Z}_{\ge 1}$) Pareto-optimal archetypes $\{(\bm{\theta}_k, \mathbf{m}_k, \ell_k)\}_{k=1}^{K_{\mathrm{eff}}}$ (Figure~\ref{fig:framework}).

\subsection{Canonical Topology and Port Recovery (CTPR)}
Raw netlists exhibit three structural failure modes: 5-node MOSFET cards carrying extra substrate tokens, floating wells, and inverted drain/source (D/S) terminals from symmetric schematic symbols. \ctpr normalizes 5-node tokens, ties floating NMOS/PMOS wells to $\mathrm{VSS}/\mathrm{VDD}$, and resolves D/S orientation prior to sizing by solving a grounded Dirichlet Laplacian system over internal nets $\mathcal{V}_{\mathrm{int}} = \mathcal{V}_{\mathrm{net}} \setminus \{\mathrm{VDD}, \mathrm{VSS}\}$ with unit channel/resistor conductances:
\begin{equation}
  \mathbf{L}_{\mathrm{ch}}\,\bm{\phi} = \mathbf{b}_{\mathrm{rail}}, \qquad \phi(\mathrm{VDD}) = 1,\;\; \phi(\mathrm{VSS}) = 0.
  \label{eq:icml_dirichlet}
\end{equation}
Swapping D and S whenever $\phi(n_{\mathrm{S}}) > \phi(n_{\mathrm{D}}) + \epsilon_\phi$ on NMOS (and $\phi(n_{\mathrm{S}}) < \phi(n_{\mathrm{D}}) - \epsilon_\phi$ on PMOS) aligns every channel with monotonic rail-to-rail dc potential drop.

\ctpr then runs 1-WL color refinement~\citep{weisfeiler1968reduction} over the pin-typed bipartite graph to compute isomorphism-invariant hash $h_{\mathrm{WL}}(\widehat{G})$ and partition MOSFETs into $D_{\mathrm{orb}}$ symmetry orbits $\{\mathcal{O}_1, \dots, \mathcal{O}_{D_{\mathrm{orb}}}\}$, compressing sizing dimensionality by \MeanOrbitCompression{}$\times$ while enforcing exact differential and mirror matching. Candidate ports in $\mathcal{P}_{\mathrm{ext}}$ are typed structurally into gate signal inputs, diode-connected bias ports, and high-impedance D/S outputs. Finally, \ctpr checks differential feedback polarity~\citep{middlebrook1975measurement,tian2001striving} and dc output common-mode stabilization~\citep{razavi2016design} in one solve:

\begin{proposition}[Single-Probe Differential Polarity and Servoed CMFB Observability]
\label{prop:single_probe}
Let $(\mathrm{in}_1, \mathrm{in}_2)$ be biased at equal dc voltage $V_{\mathrm{in}_1,\mathrm{dc}} = V_{\mathrm{in}_2,\mathrm{dc}} = V_{\mathrm{CM}} := V_{DD}/2$, with output $(\mathrm{out}_1, \mathrm{out}_2)$ ($\mathrm{out}_2 \equiv \mathrm{VSS}$ and $\mathbf{e}_{\mathrm{VSS}} := \mathbf{0} \in \R^N$ for single-ended stages, or dual non-rail outputs $\mathrm{out}_1, \mathrm{out}_2 \notin \{\mathrm{VDD}, \mathrm{VSS}\}$ tied to $V_{\mathrm{CM}}$ via matched dc-sensing termination $Z_{\mathrm{cm}}(s) = R_s/2 + sL_{\mathrm{servo}}$ with $G_{\mathrm{servo}} := 2/R_s > 0, L_{\mathrm{servo}} > 0$) converging to dc equilibrium $\mathbf{x}_0 \in \R^N$. Assume the linearized MNA matrix $\mathbf{M}(j\omega_0) = \mathbf{J}(\mathbf{x}_0) + j\omega_0 \mathbf{C}(\mathbf{x}_0)$ is nonsingular at low passband $\omega_0 > 0$ below the first $90^\circ$ phase crossover, with nonzero in-phase differential voltage gain $\Re\{A_{\mathrm{dm}}(\omega_0)\} \neq 0$ satisfying $\mathrm{sign}\,\Re\{A_{\mathrm{dm}}(\omega_0)\} = \mathrm{sign}\,\lim_{\omega \to 0^+} \Re\{A_{\mathrm{dm}}(\omega)\}$. Then one `.op` + `.ac` solve with anti-symmetric input phasor $\mathbf{u}^{(+)} = [+1/2, -1/2]^\top$ ($v_{\mathrm{in}_1} - v_{\mathrm{in}_2} = 1, v_{\mathrm{in}_1} + v_{\mathrm{in}_2} = 0$) establishes: (i)~$V_{\mathrm{out}}^{(-)}(\omega_0) = -V_{\mathrm{out}}^{(+)}(\omega_0)$ under $(\mathrm{in}_1 \leftrightarrow \mathrm{in}_2)$, so $\sigma_{\mathrm{pol}} = \mathrm{sign}\,\Re\{V_{\mathrm{out}}^{(+)}(\omega_0)\} \in \{+1, -1\}$ uniquely aligns low-frequency differential gain polarity without re-simulation; and (ii)~for dual non-rail outputs whose zero-differential-input open-loop subcircuit common-mode current $I_{\mathrm{sub}} : \R \to \R$ is $g_{\mathrm{cm}}^{\max}$-Lipschitz ($0 \le g_{\mathrm{cm}}^{\max} < G_{\mathrm{servo}}$), observing $|I_{\mathrm{servo}}(\mathbf{x}_0)| > \tau_I := \frac{g_{\mathrm{cm}}^{\max} G_{\mathrm{servo}}}{G_{\mathrm{servo}} - g_{\mathrm{cm}}^{\max}}\Delta V_{\mathrm{max}}$ ($\Delta V_{\mathrm{max}} > 0$) guarantees $I_{\mathrm{sub}}(v) \neq 0$ for all $v \in [V_{\mathrm{CM}} - \Delta V_{\mathrm{max}}, V_{\mathrm{CM}} + \Delta V_{\mathrm{max}}]$.
\end{proposition}
Proof and algorithmic disjunction remarks appear in Appendix~\ref{sec:app_proofs}.

\subsection{Hierarchical Pruning and Physics-Manifold Identification (PMCI)}
To avoid diluting simulation budgets across structurally invalid hypotheses, \pmci applies a zero-SPICE \emph{Tier-1 structural admissibility filter} $\Phi_{\mathrm{T1}}(\widehat{G}) \subseteq \mathcal{C}$ checking mandatory port counts and minimal functional subgraphs, pruning \MeanPrunedClasses{} of 10 classes per circuit. For each surviving $c \in \Phi_{\mathrm{T1}}(\widehat{G})$, all transistors in orbit $\mathcal{O}_q$ share a normalized inversion-intent vector $\bm{\theta}_q = (u_{g_m/I_D, q}, u_{I_D, q}, u_{L, q}) \in [0, 1]^3$, mapped through pre-computed 3D BSIM4 lookup tables $\mathcal{L}_{\mathrm{PDK}}(g_m/I_D, L, V_{DS})$~\citep{jespers2017systematic} to physical widths $W_q = I_{D,q} / J_D((g_m/I_D)_q, L_q)$. Batched \texttt{ngspice-42} Sobol exploration followed by separable $(\mu/\mu_W, \lambda)$-CMA-ES~\citep{hansen2001completely} maximizes worst-case primary margin $\mu_c(\bm{\theta}) = \min_{m \in \mathcal{M}_{\mathrm{prim}}(c)} \tilde{g}_m(\mathbf{m}(\bm{\theta})) / \varsigma_m$, certifying the highest-specificity class $\widehat{c}$ satisfying $\mu_{\widehat{c}}(\bm{\theta}) \ge 0$ across all primary specifications (or abstaining as \text{Unclassified}).

\subsection{Pareto-Hypervolume Trade-Off Atlas (PHTA)}
Let $\tilde{\mathbf{g}}(\mathbf{m}) \in \R^M$ ($M \in \mathbb{Z}_{\ge 1}$) denote unclipped sign-oriented linearized margins centered at thresholds $R_{\widehat{c},j}$, and let $0 < \epsilon_\sigma \le \sigma_j < \infty$ ($\epsilon_\sigma = 10^{-3}$). For penalty $\lambda \ge 1$ ($\lambda = 2.5$), \phta defines the \emph{Bi-Lipschitz Deficit-Penalized Signed Margin} $\psi_\lambda: \R^M \to \R^M$:
\begin{equation}
  \bigl[\psi_\lambda(\tilde{\mathbf{g}})\bigr]_j = (\tilde{g}_j / \sigma_j)\mathbb{I}[\tilde{g}_j \ge 0] + (\lambda\,\tilde{g}_j / \sigma_j)\mathbb{I}[\tilde{g}_j < 0].
  \label{eq:icml_psi}
\end{equation}
\phta extracts non-dominated Pareto designs $\mathcal{P} \subseteq \mathcal{F}_{\mathrm{strict}}$ when $\mathcal{F}_{\mathrm{strict}} = \{i \in \mathcal{F} : \tilde{\mathbf{g}}(\mathbf{m}_i) \ge \mathbf{0}\} \neq \emptyset$, falling back to primary-feasible $\mathcal{F}$ when $\mathcal{F}_{\mathrm{strict}} = \emptyset$ (activating Eq.~\eqref{eq:icml_psi}'s $\lambda > 1$ deficit penalty). Over $\mathcal{Y} = \{\psi_\lambda(\tilde{\mathbf{g}}(\mathbf{m}_i)) : i \in \mathcal{P}\}$ ($1 \le |\mathcal{Y}| < \infty$), let $\mathbf{r}_{\mathrm{ref}} := \min_{\mathbf{y} \in \mathcal{Y}} \mathbf{y} - \delta_{\mathrm{pad}}\mathbf{1}$ ($\delta_{\mathrm{pad}} = 0.25 > 0$) and $\mathbf{u}_{\mathrm{top}} := \max_{\mathbf{y} \in \mathcal{Y}} \mathbf{y} + \epsilon_{\mathrm{top}}\mathbf{1}$ ($\epsilon_{\mathrm{top}} = 10^{-6} > 0$), and draw $Q = 4096$ ($Q \in \mathbb{Z}_{\ge 1}$) scrambled Sobol points $\{\mathbf{q}_r\}_{r=1}^Q \subset [\mathbf{r}_{\mathrm{ref}}, \mathbf{u}_{\mathrm{top}}]$. \phta maximizes dominated set-coverage hypervolume:
\begin{equation}
  F_{\mathrm{HV}}(S) = w \sum_{r=1}^Q \mathbb{I}\!\left[\exists\,\mathbf{y} \in S \text{ s.t. } \mathbf{q}_r \le \mathbf{y}\right], \quad w := \frac{\mathrm{Vol}([\mathbf{r}_{\mathrm{ref}}, \mathbf{u}_{\mathrm{top}}])}{Q} > 0.
  \label{eq:icml_hv}
\end{equation}

\begin{proposition}[Bi-Lipschitz Pareto Fidelity and Submodular Coverage Bound]
\label{prop:phta}
Let $\lambda \ge 1$, $0 < \epsilon_\sigma \le \sigma_j < \infty$ ($j=1,\dots,M$), $1 \le |\mathcal{Y}| < \infty$, $K, Q, M \in \mathbb{Z}_{\ge 1}$, and $\tilde{\mathbf{g}} \succ \tilde{\mathbf{g}}'$ denote strict Pareto dominance ($\tilde{g}_j \ge \tilde{g}'_j\;\forall j$ and $\exists j^* : \tilde{g}_{j^*} > \tilde{g}'_{j^*}$). Then:
(a)~$\psi_\lambda: \R^M \to \R^M$ is a homeomorphism with exact inverse $[\psi_\lambda^{-1}(\mathbf{y})]_j = \sigma_j y_j \mathbb{I}[y_j \ge 0] + (\sigma_j / \lambda)y_j \mathbb{I}[y_j < 0]$, tight global bi-Lipschitz constants $L_{\min} = \min_j \sigma_j^{-1} > 0$ and $L_{\max} = \lambda \max_j \sigma_j^{-1} \le \lambda/\epsilon_\sigma < \infty$, and strict Pareto order preservation ($\tilde{\mathbf{g}} \succ \tilde{\mathbf{g}}' \iff \psi_\lambda(\tilde{\mathbf{g}}) \succ \psi_\lambda(\tilde{\mathbf{g}}')$); and
(b)~$F_{\mathrm{HV}}$ is normalized, monotone, and submodular, so greedy maximization terminates in $K_{\mathrm{eff}} = \min(K, |\mathcal{Y}|)$ steps with $F_{\mathrm{HV}}(S_{\mathrm{greedy}}) \ge (1 - 1/e)\max_{|S| \le K} F_{\mathrm{HV}}(S)$ in $O(|\mathcal{Y}| Q (M + \min(K, |\mathcal{Y}|))) \subseteq O(K |\mathcal{Y}| Q M)$ time.
\end{proposition}
Proofs appear in Appendix~\ref{sec:app_proofs}. Each selected archetype receives a dual-sided tag $\ell_k = \texttt{<MaxSurplus>\_vs\_<MinMarginCost>-A}k$.

\subsection{Unsupervised Curation, PVT Certification, and Transfer}
\strata requires zero gradient training during corpus curation, optionally certifying \texttt{SKY130} amplifier archetypes across 15/45 PVT corners and 50-trial Pelgrom mismatch~\citep{pelgrom1989matching}. At inference on an unseen topology $G_{\mathrm{new}}$ ($h_{\mathrm{WL}}(\widehat{G}_{\mathrm{new}}) \notin \mathcal{D}_{\mathrm{atlas}}$), standardized 1-WL $k$-NN retrieves top-3 atlas neighbors and projects their diverse archetype vectors $\bm{\theta}^*$ onto role-matched orbits of $\widehat{G}_{\mathrm{new}}$ to warm-start separable CMA-ES sizing or transfer across CMOS nodes (`130nm <-> 45nm <-> 28nm`).

\section{EXPERIMENTS}
\label{sec:experiments}

\subsection{Experimental Setup}
We evaluate \strata across \TotalRawCircuits{} public circuit graphs combining \texttt{AMSNet~1.0} (\AMSNetRawCount{} schematics/netlists)~\citep{tao2024amsnet} and \texttt{AnalogGenie} (\AnalogGenieRawCount{} human-categorized subcircuits)~\citep{gao2025analoggenie}, totaling \TotalSpiceEvals{} unique BSIM4v4.7 evaluations across \TotalNgspiceProcs{} batched \texttt{ngspice-42} processes on \texttt{SkyWater SKY130} 130nm ($1.8\;\mathrm{V}$)~\citep{skywater2020pdk}, \texttt{PTM} 45nm ($1.1\;\mathrm{V}$), and \texttt{PTM} 28nm ($1.0\;\mathrm{V}$)~\citep{zhao2006new}. Ground-truth provenance is explicitly partitioned into:
(1)~pure-human \texttt{real\_gt} ($N=\HumanN{}$ non-isomorphic \texttt{AnalogGenie} graphs assigned directly from author chapter ranges and \texttt{Port\#.txt} output counts with zero \ctpr calls; cross-corpus \texttt{AMSNet-78} with zero \texttt{Port\#.txt}; and unmodified \texttt{AMSNet-734} \texttt{.cir} connectivity);
(2)~structurally audited proxy benchmarks ($^\dagger$\texttt{self\_supervised\_proxy}: \texttt{AMSNet-734} and \texttt{Stratified-352}, where incomplete 1--2 transistor textbook fragments are audited as \text{Unclassified}); and
(3)~SPICE physical benchmarks (\texttt{simulation\_only}). Blind sanitization strips all identifiers and permutes node and device order before inference (enforced via automated AST zero-leakage and permutation-invariance unit tests).

\begin{table*}[t]
  \centering
  \caption{\textbf{Unsupervised 11-Class Circuit Identification Across Pure-Human \texttt{real\_gt} and Structurally Audited ($^\dagger$\texttt{self\_supervised\_proxy}) Benchmarks.} Top rows show 0-SPICE structural/supervised references without physical simulation verification; bottom rows compare blind SPICE-verified unsupervised pipelines given unlabeled boundary pins ($B=64$ on \texttt{Pure-Human-330}/\texttt{Strat-352}, $B=48$ on \texttt{AMSNet-78}/\texttt{734}).}
  \label{tab:identification}
  \resizebox{\textwidth}{!}{\begin{tabular}{lccccccc}
\toprule
 & \multicolumn{3}{c}{\textbf{Pure-Human \texttt{real\_gt} ($N=330$)}} & \multicolumn{2}{c}{\textbf{Cross-Corpus ($N=78$)}} & \textbf{AMSNet-734} & \textbf{Strat-352} \\
\cmidrule(lr){2-4} \cmidrule(lr){5-6} \cmidrule(lr){7-7} \cmidrule(lr){8-8}
\textbf{Method / Paradigm} & \textbf{Acc.} & \textbf{Macro-F1} & \textbf{SPICE} & \textbf{Acc.} & \textbf{Macro-F1} & \textbf{Macro-F1}$^\dagger$ & \textbf{Macro-F1}$^\dagger$ \\
\midrule
\multicolumn{8}{l}{\textit{Zero-SPICE Structural / Supervised Graph Reference (No Simulation Verification)}} \\
WL-RGCN (5-fold CV, supervised) & 0.600 & 0.597 & 0 & --- & --- & 0.274 & 0.586 \\
Rule-Subgraph (0-SPICE heuristic) & 0.509 & 0.468 & 0 & 0.513 & 0.175 & 0.480 & 0.649 \\
\midrule
\multicolumn{8}{l}{\textit{Blind Unsupervised SPICE-Verified Circuit Identification (Requires Physical Sizing Closure)}} \\
AMSnet-q-Repro (28nm, $1\times B$) & 0.136 & 0.054 & 16,352 & 0.179 & 0.052 & 0.107 & 0.061 \\
AMSnet-q-Repro (SKY130 130nm, $1\times B$) & --- & --- & --- & --- & --- & --- & 0.099 \\
AMSnet-q + ERC (28nm, $1\times B$) & --- & --- & --- & --- & --- & 0.109 & 0.072 \\
AMSnet-q-Enhanced (28nm, $2.25\times B$+ERC+CMFB) & 0.158 & 0.065 & 34,758 & 0.064 & 0.022 & --- & --- \\
\textbf{STRATA (SKY130 130nm, $1\times B$)} & 0.212 & 0.215 & 12,924 & 0.359 & \textbf{0.324} & \textbf{0.475} & \textbf{0.479} \\
\textbf{STRATA (PTM 28nm, $1\times B$)} & \textbf{0.273} & \textbf{0.302} & 16,078 & \textbf{0.372} & 0.294 & 0.383 & 0.443 \\
\bottomrule
\end{tabular}
}
\end{table*}

\subsection{Main Quantitative Comparisons}
\noindent\textbf{Unsupervised Circuit Identification.}
To evaluate \strata's ability to identify functional circuit classes while enforcing closed-loop SPICE feasibility, Table~\ref{tab:identification} compares all baselines across four benchmarks. On pure-human \texttt{real\_gt} ($N=\HumanN{}$), \strata (28nm, $B=64$) achieves \StrataHumanFOne{} Macro-F1 (\StrataHumanAcc{} accuracy), outperforming matched-28nm \texttt{AMSnet-q-Repro} (\AmsqReproHumanFOne{}) by \HumanGainOverReproX{}$\times$ and \texttt{AMSnet-q-Enhanced} (\AmsqEnhHumanFOne{}) by \HumanGainOverEnhX{}$\times$ using $2.16\times$ fewer SPICE evaluations (\StrataHumanEvals{} vs.\ \AmsqEnhHumanEvals{}). Zero-SPICE \texttt{Rule-Subgraph} (\RuleHumanFOne{}) and supervised \texttt{WL-RGCN} (\WlRgcnHumanFOne{}) obtain higher scores on raw chapter labels because they label partial textbook sub-sketches without checking SPICE specification closure, whereas \strata abstains when physical sizing fails. On cross-corpus \texttt{AMSNet-78} (\texttt{real\_gt}, zero \texttt{Port\#.txt}), matched-28nm \strata reaches \CrossAmsPtmFOneThree{} vs.\ \texttt{AMSnet-q-Repro}'s $0.052$ (\CrossAmsSamePdkGainX{}$\times$), and \strata (\texttt{SKY130}) reaches \CrossAmsSkyFOneThree{} (\CrossAmsGainX{}$\times$ vs.\ 28nm \texttt{AMSnet-q}). On \texttt{AMSNet-734}$^\dagger$ and \texttt{Stratified-352}$^\dagger$, \strata (\texttt{SKY130}) reaches \AmsFullStrataSkyFOne{} (\AmsFourWayStrataSkyFOne{} 4-way F1 vs.\ \AmsFourWayQReproFOne{}, verifying \AmsVerStrataSky{} complete topologies) and \StratStrataSkyFOne{} Macro-F1 (\StratSameSkyGainX{}$\times$ over matched-\texttt{SKY130} \texttt{AMSnet-q-Repro} at \StratQReproSkyFOne{}, while matched-28nm \strata beats \texttt{AMSnet-q + ERC} by \StratSamePtmGainX{}$\times$), with \PortRoleAccSky{}\% port-role accuracy and \PolarityConsistPct{}\% single-probe polarity consistency ($N=74$).

\begin{table}[t]
  \centering
  \caption{\textbf{Trade-Off Labeling Fidelity ($N=32$ \texttt{SKY130} Pools, 1,000 Queries) and Multi-Corner PVT / Pelgrom Yield ($N=18$ Amplifiers, \texttt{simulation\_only}).}}
  \label{tab:phta_pvt}
  \resizebox{\columnwidth}{!}{\begin{tabular}{lccccc}
\toprule
\textbf{Archetype Labeling Method ($K=5$)} & \textbf{FFR (\% $\downarrow$)} & \textbf{HDVR (\% $\downarrow$)} & \textbf{NHV ($\uparrow$)} & \textbf{Spec P@1 (\% $\uparrow$)} & \textbf{Spec P@5 (\% $\uparrow$)} \\
\midrule
AMSnet-q Alg.\ 2 ($\max(\mathbf{Z},0)$ $K$-Means) & 13.75 & 16.25 & 0.1519 & 18.3 & 59.5 \\
Standard $Z$-Score $K$-Means & 15.00 & 15.62 & 0.1368 & 20.8 & 56.1 \\
GMM Medoids & 15.00 & 18.75 & 0.1237 & 20.8 & 52.3 \\
Marginal Quantile Tagging & 3.12 & 15.62 & 0.1546 & 14.5 & 43.0 \\
Pareto-Only Euclidean $K$-Means & \textbf{0.00} & \textbf{3.12} & 0.1646 & 28.6 & 68.7 \\
\textbf{STRATA-PHTA ($\psi_\lambda$ + QMC Submodular)} & \textbf{0.00} & \textbf{3.12} & \textbf{0.2075} & \textbf{40.0} & \textbf{78.2} \\
\midrule
\multicolumn{6}{l}{\textit{SKY130 130nm Multi-Corner PVT \& 50-Trial Pelgrom Mismatch Certification ($N=18$ Amplifiers, $B=84$)}} \\
\textbf{Pipeline Setting} & \textbf{15-Corner PVT} & \textbf{45-Corner PVT} & \textbf{50-MC Yield} & \textbf{Mean $V_{\mathrm{os},3\sigma}$} & \textbf{Conv.} \\
\midrule
AMSnet-q End-to-End (Matched $B=84$ + ERC + CMFB) & 5.56\% & 5.56\% & 0.00\% & --- & 5/18 \\
AMSnet-q Alg.\ 2 (Given STRATA SKY130 Pool) & 80.00\% & 78.77\% & \textbf{83.33\%} & \textbf{17.17\,mV} & 18/18 \\
\textbf{STRATA-PHTA (Full Pipeline, $B=84$)} & \textbf{84.07\%} & \textbf{83.09\%} & \textbf{83.33\%} & 26.06\,mV & \textbf{18/18} \\
\bottomrule
\end{tabular}
}
\end{table}

\noindent\textbf{Schematic Self-Healing and \phta Pareto Fidelity.}
To evaluate \ctpr on noisy schematic graphs, Table~\ref{tab:vision_ablation} (top) tests all \AMSNetRawCount{} \texttt{AMSNet~1.0} schematics against unmodified \texttt{.cir} files (\texttt{real\_gt}): \texttt{STRATA-Vision + CTPR} raises All-734 Netlist-F1 from \VisRawBfsFOne{} (\texttt{AMSNet} BFS) and \VisRawSinaFOne{} (\texttt{SINA} CCL) to \VisRawStrataCtprFOne{}, and recovers \HealCtprCompilePct{}\% SPICE compilation (vs.\ \HealRawCompilePct{}\%), \HealCtprFuncPct{}\% AC equivalence (vs.\ \HealRawFuncPct{}\%), and \HealDsOrientPct{}\% D/S orientation on $N=95$ corrupted graphs. In Table~\ref{tab:phta_pvt}, strict Pareto filtering ($\mathcal{P}\subseteq\mathcal{F}_{\mathrm{strict}}$, matched on FFR/HDVR by \texttt{Pareto-Only Euclidean K-Means}) eliminates \texttt{AMSnet-q}'s \AmsqFFR{}\% FFR (\PhtaFFR{}\%) and lowers HDVR to \PhtaHDVR{}\% (where $\psi_\lambda$'s $\lambda$-branch governs partial-feasibility fallback), while Proposition~\ref{prop:phta}b's submodular hypervolume selection drives a $+26.1\%$ NHV gain (\PhtaNHV{} vs.\ \ParetoKmNHV{}) and lifts $k=1$ query Precision@1 from \ParetoKmPOne{}\% to \PhtaPOne{}\% ($2.19\times$ over \texttt{AMSnet-q}'s \AmsqPOne{}\%). Across $N=18$ \texttt{SKY130} amplifiers, \strata achieves \PhtaPvtFifteen{}\% 15-corner PVT (\PhtaPvtFortyFive{}\% 45-corner) and \PhtaPelgromYield{}\% Pelgrom yield ($18/18$ converged vs.\ $5/18$ and \EteAmsqPvtFifteen{}\% for end-to-end \texttt{AMSnet-q}), beating same-pool \texttt{AMSnet-q} by $+4.07/+4.32$ PVT points while trading higher mean offset across minimum-area Pareto corners (\PhtaVosMv{} vs.\ \SamePoolAmsqVosMv{}\;mV) at identical \PhtaPelgromYield{}\% yield.

\begin{figure}[t]
  \centering
  \includegraphics[width=\columnwidth]{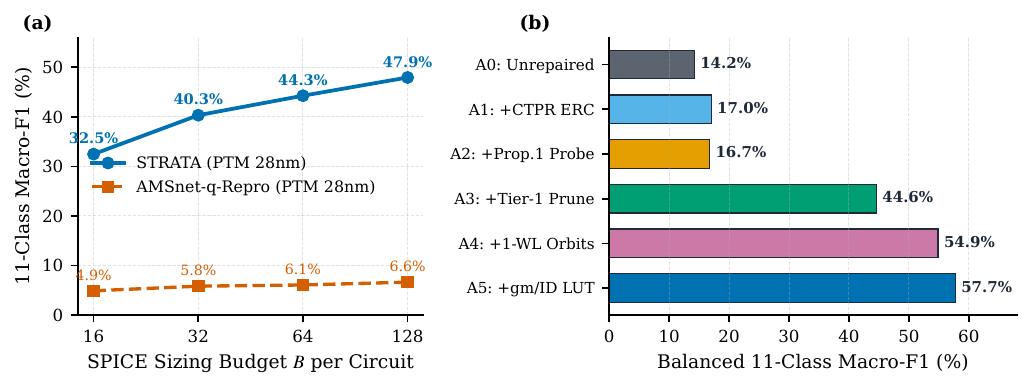}
  \caption{\textbf{Budget Scaling and Component Ablation.} (a)~Macro-F1 across SPICE budgets $B\in\{16,32,64,128\}$ on 28nm \texttt{Stratified-352}$^\dagger$. (b)~Complete cumulative ablation $A_0 \dots A_5$ ($N=110$, $B=56$).}
  \label{fig:scaling_ablation}
\end{figure}

\begin{table}[t]
  \centering
  \caption{\textbf{AMSNet-734 Schematic Extraction (\texttt{real\_gt}) and Complete Cumulative Component Ablation ($N=110$, \texttt{self\_supervised\_proxy}).}}
  \label{tab:vision_ablation}
  \resizebox{\columnwidth}{!}{\begin{tabular}{lcccc}
\toprule
\textbf{AMSNet-734 Schematic Extractor (\texttt{real\_gt})} & \textbf{Easy F1} & \textbf{Med.\ F1} & \textbf{Hard F1} & \textbf{All-734 Netlist-F1 (Conn-F1)} \\
\midrule
AMSNet-1.0 Pixel BFS & 0.4727 & 0.4516 & 0.4185 & 0.4467 (0.2864) \\
SINA-Style Connected-Component CCL & 0.4764 & 0.4500 & 0.4149 & 0.4451 (0.2843) \\
STRATA-Vision (Raw Junction Solver) & 0.4809 & 0.4556 & 0.4212 & 0.4507 (0.2919) \\
\textbf{STRATA-Vision + CTPR Self-Healing} & \textbf{0.5134} & \textbf{0.4986} & \textbf{0.4496} & \textbf{0.4902 (0.3450)} \\
\midrule
\textbf{Cumulative Ablation ($N=110$, $B=56$, Proxy)} & \multicolumn{2}{c}{\textbf{Macro-F1 ($\uparrow$)}} & \textbf{Accuracy ($\uparrow$)} & \textbf{Mean SPICE / Ckt} \\
\midrule
$A_0$: Raw Unrepaired Per-MOS Search & \multicolumn{2}{c}{0.1420} & 0.2000 & 48.5 \\
$A_1$: $+\,$CTPR Syntax \& D/S Orientation Repair & \multicolumn{2}{c}{0.1703} & 0.2273 & 48.7 \\
$A_2$: $+\,$Prop.\ 1 Single-Probe \& CMFB Check & \multicolumn{2}{c}{0.1674} & 0.2273 & 48.1 \\
$A_3$: $+\,$Tier-1 Structural Admissibility Pruning & \multicolumn{2}{c}{0.4458} & 0.4091 & 49.8 \\
$A_4$: $+\,$WL-1 Symmetry-Orbit Tying ($1.89\times$) & \multicolumn{2}{c}{0.5486} & 0.4909 & 49.3 \\
\textbf{$A_5$: Full STRATA ($+\,g_m/I_D$ BSIM4 LUT)} & \multicolumn{2}{c}{\textbf{0.5774}} & \textbf{0.5182} & 48.9 \\
\bottomrule
\end{tabular}
}
\end{table}

\subsection{Ablation and Hyperparameter Sensitivity}
\begin{figure}[t]
  \centering
  \includegraphics[width=\columnwidth]{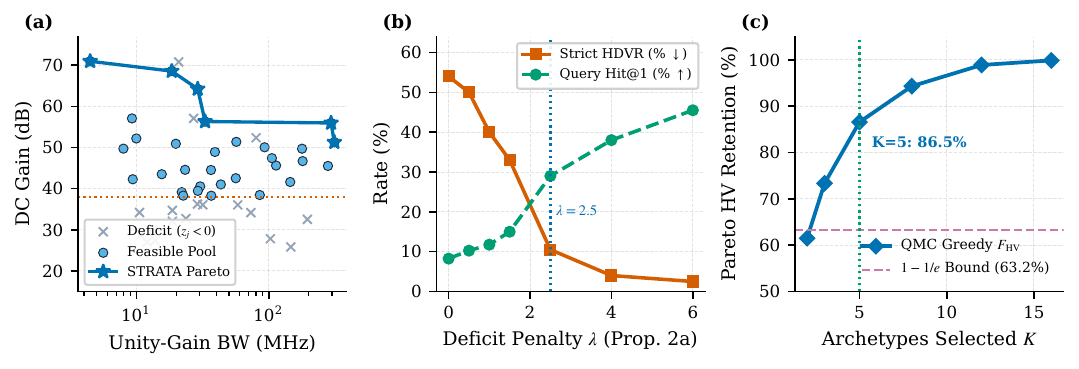}
  \caption{\textbf{\phta Geometry and Sensitivity.} (a)~Real \texttt{SKY130} \texttt{SE-OpAmp} Pareto frontier vs.\ deficit points. (b)~Un-filtered pool $\psi_\lambda$ sweep. (c)~Pareto HV retention vs.\ $K$.}
  \label{fig:phta_sens}
\end{figure}

Figure~\ref{fig:scaling_ablation}, Table~\ref{tab:vision_ablation} (bottom), and Figure~\ref{fig:phta_sens} evaluate budget scaling, cumulative ablations ($N=110$, $B=56$), and $\lambda/K$ sweeps. Across $B \in \{16,32,64,128\}$ on 28nm, \strata scales from \BudgetScaleMinPct{}\% to \BudgetScaleMaxPct{}\% Macro-F1 whereas \texttt{AMSnet-q-Repro} plateaus at $6.6\%$. Tracing $A_0 \to A_5$: \ctpr repair lifts F1 from \AblBaseFOne{} to \AblErcFOne{} ($A_1$), pre-pruning single-probe checks give \AblProbeFOne{} ($A_2$), Tier-1 structural pruning provides the primary jump to \AblTierOneFOne{} ($A_3$, $+0.2784$ F1; dropping to \AblLooNoTierOneFOne{} in LOO due to $10\times$ budget dilution), 1-WL symmetry orbits add $+0.1028$ to \AblOrbitsFOne{} ($A_4$), and BSIM4 $g_m/I_D$ LUTs reach \AblFullFOne{} ($A_5$; complete 11-row LOO matrix in Appendix~\ref{sec:app_specs}). Disabling $\mathcal{F}_{\mathrm{strict}}$ pre-filtering in Figure~\ref{fig:phta_sens}b isolates $\psi_\lambda$ on mixed pools, cutting strict HDVR from \LamZeroStrictHdvr{}\% ($\lambda=0$ clipping) and \LamOneStrictHdvr{}\% ($\lambda=1$) to \LamStdStrictHdvr{}\% ($\lambda=2.5$, $\mathrm{NHV}=\LamStdUnfiltNHV{}$) and \LamSixStrictHdvr{}\% ($\lambda=6.0$), while $K=5$ retains \KFiveHvRetain{}\% of full Pareto hypervolume (Figure~\ref{fig:phta_sens}c).

\begin{figure}[t]
  \centering
  \includegraphics[width=\columnwidth]{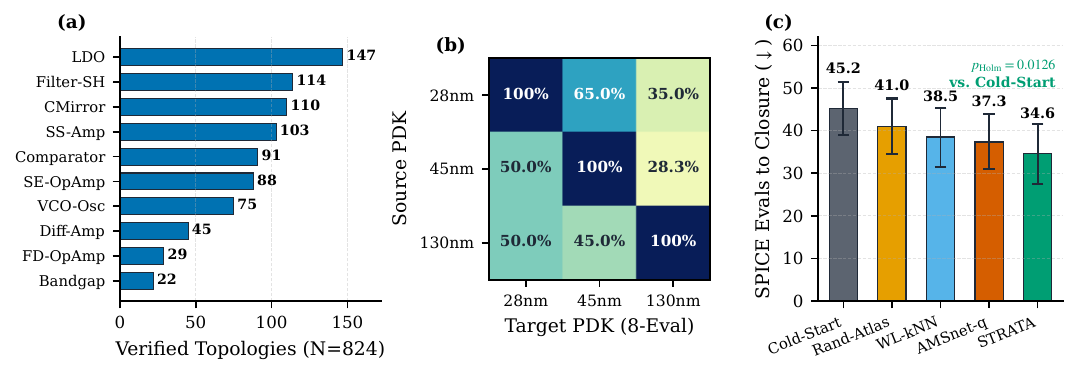}
  \caption{\textbf{\strata-Atlas Scale, Cross-PDK Transfer, and Disjoint Sizing.} (a)~10-class verified topologies ($N=\VerifiedAtlasTopologies{}$). (b)~8-eval cross-PDK retention ($N=60$). (c)~5-fold disjoint OpAmp sizing ($N=50$).}
  \label{fig:atlas_downstream}
\end{figure}

\begin{table}[t]
  \centering
  \caption{\textbf{5-Fold Strictly Topology-Disjoint Downstream Sizing (\texttt{simulation\_only}) and Blind-Graph Envelope Prediction (\texttt{self\_supervised\_proxy}; $^\S$oracle given \pmci's feasible pool).}}
  \label{tab:downstream}
  \resizebox{\columnwidth}{!}{\begin{tabular}{lccccc}
\toprule
\textbf{Task 6A: Disjoint OpAmp Sizing ($N=50$, 5-Fold)} & \textbf{Mean Evals ($\downarrow$)} & \textbf{95\% Boot.\ CI} & \textbf{Speedup} & \textbf{$p_{\mathrm{Holm}}$ vs.\ STRATA} & \textbf{Closure (\% $\uparrow$)} \\
\midrule
Cold-Start Separable CMA-ES & 45.16 & [39.0, 51.4] & $1.00\times$ & 0.0126 & 50.0\% \\
Random-Atlas Warm-Start ($k=3$) & 41.04 & [34.6, 47.5] & $1.10\times$ & 0.176 & 60.0\% \\
WL-Graph $k$-NN Nominal Init ($k=3$) & 38.52 & [31.5, 45.3] & $1.17\times$ & 0.238 & 58.0\% \\
AMSnet-q Medoid $k$-NN Init ($k=3$) & 37.32 & [31.0, 43.9] & $1.21\times$ & 0.238 & 64.0\% \\
\textbf{STRATA-PHTA Frontier $k$-NN ($k=3$)} & \textbf{34.60} & \textbf{[27.6, 41.6]} & \textbf{$1.31\times$} & --- & \textbf{66.0\%} \\
\midrule
\textbf{Task 6B: Blind-Graph Envelope ($N=265$, Proxy)} & \multicolumn{3}{c}{\textbf{Shared Feasible Ref $R^2$ ($\uparrow$)}} & \multicolumn{2}{c}{\textbf{Gain MAE (dB $\downarrow$)}} \\
\midrule
Single-Sizing Nominal Labeler ($k=1$) & \multicolumn{3}{c}{$-0.7936 \pm 0.1450$} & \multicolumn{2}{c}{$12.89 \pm 1.81$} \\
AMSnet-q Alg.\ 2 Medoid Envelope ($k=5$, Unfiltered) & \multicolumn{3}{c}{$0.3587 \pm 0.0259$} & \multicolumn{2}{c}{$4.79 \pm 0.39$} \\
AMSnet-q Alg.\ 2 (Given PMCI Feasible Pool, $k=5$)$^\S$ & \multicolumn{3}{c}{\textbf{$0.4416 \pm 0.0566$}} & \multicolumn{2}{c}{\textbf{$4.77 \pm 0.36$}} \\
\textbf{STRATA-PHTA Archetype Envelope ($k=5$, End-to-End)} & \multicolumn{3}{c}{$0.4041 \pm 0.0721$} & \multicolumn{2}{c}{$4.80 \pm 0.50$} \\
\bottomrule
\end{tabular}
}
\end{table}

\subsection{Atlas Scale, Cross-PDK Transfer, and Downstream Utility}
Processing all \TotalRawCircuits{} graphs (Figure~\ref{fig:atlas_downstream}, Table~\ref{tab:downstream}) removes \DuplicatesRemoved{} 1-WL duplicates (\CrossCorpusOverlaps{} cross-corpus overlaps), yielding \UniqueWLTopologies{} unique topologies and \VerifiedAtlasTopologies{} verified 28nm topologies (\TotalAtlasArchetypes{} archetypes) plus \SkyCohortVerified{} \texttt{SKY130} and \PtmFortyFiveCohortVerified{} 45nm cards (\MultiPDKTotalCards{} multi-PDK cards), with \XferTwentyeightToFortyfiveZS{}\% zero-shot and \XferTwentyeightToFortyfiveWarm{}\% 8-eval $28\mathrm{nm}\to 45\mathrm{nm}$ transfer retention. On 5-fold topology-disjoint OpAmp sizing ($N=50$), \strata-\phta warm-start closes specs in \WarmStrataEvals{} evaluations (\WarmStrataClosure{}\% closure, \WarmStrataSpeedup{}$\times$ over cold-start separable CMA-ES at \WarmColdEvals{} evaluations, $p_{\mathrm{Holm}}=\WarmWilcoxonP{}$), with directional mean reductions over warm-start baselines ($p_{\mathrm{Holm}}\in[\WarmRandP{},\WarmAmsqP{}]$). On 5-fold disjoint blind-graph envelope regression ($N=265$ amplifiers, pure parser WL features), $k=1$ single sizing collapses ($R^2=\EnvSingleRTwo{}$), end-to-end \strata ($k=5$) improves over unfiltered \texttt{AMSnet-q} ($R^2=\EnvStrataRTwo{}$ vs.\ \EnvAmsqUnfiltRTwo{}, and $R^2=\EnvFullAtlasRTwo{}$ across $N=\VerifiedAtlasTopologies{}$), and oracle feasible-pool \texttt{AMSnet-q}$^\S$ reaches \EnvAmsqFeasRTwo{}, confirming \pmci feasibility filtering drives envelope predictability.

\subsection{Qualitative Analysis and Scope Boundaries}
On telescopic differential OTA \texttt{Book281} (\texttt{ag\_281}, 8 MOSFETs in 4 orbits; Figure~\ref{fig:framework}), \phta extracts a 40-point \texttt{SKY130} Pareto frontier spanning \texttt{Compact-Area-Minimal\_vs\_ModerateSlew-A1} ($15.68\;\mu\mathrm{m}^2$, $58.98\;\mathrm{dB}$), \texttt{Wideband-High-Speed\_vs\_ModerateSlew-A3} ($68.65\;\mathrm{dB}$, $104.41\;\mathrm{MHz}$), and \texttt{Ultra-Low-Power\_vs\_ModerateSlew-A4} ($1.33\;\mu\mathrm{W}$), while certified \texttt{ag\_155} archetypes achieve $100.0\%$ 15-corner PVT survival and $100.0\%$ Pelgrom yield. Appendix~\ref{sec:app_forensics} details all boundary forensics, including \texttt{STRATA-Portless}, fixed-budget probe ablations, and parasitic sweeps ($C_{\mathrm{par}}\le 100\;\mathrm{fF}$, \CparHundredRetain{}\% retention).

\section{CONCLUSION}
\label{sec:conclusion}

We presented \strata, an unsupervised physics-verified graph discovery and labeling framework that resolves verification-blind corpus collapse when converting raw analog circuit repositories into multi-PDK performance atlases. By coupling \ctpr's Laplacian potential orientation, 1-WL symmetry-orbit compression, and single-probe polarity/CMFB observability with \pmci's Tier-1 admissibility pruning and BSIM4 $g_m/I_D$ lookup sizing, and \phta's strict Pareto filtering, bi-Lipschitz deficit fallback, and submodular hypervolume selection, \strata bridges discrete graph automorphisms and continuous BSIM4 device physics. Across \TotalRawCircuits{} public circuit graphs and \TotalSpiceEvals{} \texttt{ngspice-42} evaluations on 130nm, 45nm, and 28nm CMOS, \strata boosts blind simulation-verified 11-class Macro-F1 by \HumanGainOverEnhX{}--\StratSamePtmGainX{}$\times$ on matched 28nm CMOS (\StratSameSkyGainX{}$\times$ on matched \texttt{SKY130}), eliminates false-frontier medoids while doubling $k=1$ specification retrieval precision, and releases \VerifiedAtlasTopologies{} verified topologies (\MultiPDKTotalCards{} multi-PDK cards) that accelerate disjoint OpAmp sizing by \WarmStrataSpeedup{}$\times$ over cold-start optimization ($p_{\mathrm{Holm}}=\WarmWilcoxonP{}$). Because \ctpr depends on exposed unlabeled boundary terminals when subcircuits omit degree-1 schematic ports and models schematic-stage lumped wire capacitances rather than post-route $RC$ trees, performance remains bounded on port-stripped macros and interconnect-limited GHz blocks. Future work will integrate automated analog layout generators and parasitic extraction~\citep{kunal2019align,xu2019magical} directly into \strata-Atlas curation.

\newpage
\section*{Impact Statement}
This paper presents research whose goal is to advance machine learning and electronic design automation through open, simulation-verified circuit datasets and multi-objective graph labeling algorithms. By curating public textbook schematics and academic SPICE netlists across open-source and predictive CMOS process models (\texttt{SkyWater SKY130}, \texttt{PTM} 45nm, and \texttt{PTM} 28nm), \strata lowers the compute and proprietary-data barriers for academic research on energy-efficient analog sensor interfaces and power management circuits. All circuit benchmarks are strictly restricted to benign, public-domain analog building blocks and open-source simulation tools (\texttt{ngspice-42}), and we identify no foreseeable negative societal or dual-use risks beyond standard integrated-circuit design automation.

\bibliography{refs}
\bibliographystyle{icml2027}

\newpage
\appendix
\onecolumn
\renewcommand{\thetable}{A\arabic{table}}
\renewcommand{\thefigure}{A\arabic{figure}}
\renewcommand{\theequation}{A\arabic{equation}}
\setcounter{table}{0}
\setcounter{figure}{0}
\setcounter{equation}{0}
\section{Mathematical Proofs and Obligations}
\label{sec:app_proofs}

\subsection{Proof of Proposition 1 (Single-Probe Differential Polarity and Servoed CMFB Observability)}
\noindent\textbf{Explicit Hypotheses.}
Let $\widehat{G}$ be a repaired circuit graph with differential input pair $(\mathrm{in}_1, \mathrm{in}_2)$ and output terminal(s) $(\mathrm{out}_1, \mathrm{out}_2)$, using $\mathbf{e}_{\mathrm{VSS}} := \mathbf{0} \in \R^N$ when $\mathrm{out}_2 \equiv \mathrm{VSS}$. Assume:
(H1)~Both inputs share equal dc common-mode bias $V_{\mathrm{in}_1,\mathrm{dc}} = V_{\mathrm{in}_2,\mathrm{dc}} = V_{\mathrm{CM}} := V_{DD}/2$, and for dual non-rail outputs ($\mathrm{out}_1, \mathrm{out}_2 \notin \{\mathrm{VDD}, \mathrm{VSS}\}$), the output midpoint connects to $V_{\mathrm{CM}}$ through matched dc-sensing termination $Z_{\mathrm{cm}}(s) = R_s/2 + sL_{\mathrm{servo}}$ ($G_{\mathrm{servo}} := 2/R_s > 0$, $L_{\mathrm{servo}} > 0$), converging in `.op` to a finite dc equilibrium $\mathbf{x}_0 \in \R^N$;
(H2)~At $\mathbf{x}_0$ and low passband $\omega_0 > 0$, the linearized Modified Nodal Analysis (MNA) matrix $\mathbf{M}(j\omega_0) = \mathbf{J}(\mathbf{x}_0) + j\omega_0 \mathbf{C}(\mathbf{x}_0) \in \mathbb{C}^{N \times N}$ is nonsingular; and
(H3)~$\omega_0$ lies below the first $90^\circ$ phase crossover with nonzero in-phase differential voltage gain $\Re\{A_{\mathrm{dm}}(\omega_0)\} \neq 0$ satisfying $\mathrm{sign}\,\Re\{A_{\mathrm{dm}}(\omega_0)\} = \mathrm{sign}\,\lim_{\omega \to 0^+} \Re\{A_{\mathrm{dm}}(\omega)\}$.

\begin{proof}
Around $\mathbf{x}_0$, small-signal nodal phasors $\mathbf{v}(\omega_0) \in \mathbb{C}^N$ satisfy $\mathbf{M}(j\omega_0)\mathbf{v}(\omega_0) = \mathbf{B}_{\mathrm{in}}\mathbf{u}_{\mathrm{in}}$, where $\mathbf{B}_{\mathrm{in}} \in \R^{N \times 2}$ maps input phasors $\mathbf{u}_{\mathrm{in}} = [v_{\mathrm{in}_1}, v_{\mathrm{in}_2}]^\top$ into MNA excitations and $V_{\mathrm{out}}(\omega_0) = (\mathbf{e}_{\mathrm{out}_1} - \mathbf{e}_{\mathrm{out}_2})^\top \mathbf{v}(\omega_0)$.

\emph{(i) Single-probe differential voltage-gain polarity.}
Exciting anti-symmetric $\mathbf{u}^{(+)} = [+1/2, -1/2]^\top$ ($v_{\mathrm{in}_1} - v_{\mathrm{in}_2} = 1$ and $v_{\mathrm{in}_1} + v_{\mathrm{in}_2} = 0$, injecting zero common-mode input even when $A_{\mathrm{cm}}(\omega_0) \neq 0$) yields $V_{\mathrm{out}}^{(+)}(\omega_0) = A_{\mathrm{dm}}(\omega_0)$ by (H2). Because $V_{\mathrm{in}_1,\mathrm{dc}} = V_{\mathrm{in}_2,\mathrm{dc}} = V_{\mathrm{CM}}$ by (H1), swapping $(\mathrm{in}_1 \leftrightarrow \mathrm{in}_2)$ preserves $\mathbf{x}_0$ and $\mathbf{M}(j\omega_0)$ while applying $\mathbf{u}^{(-)} = -\mathbf{u}^{(+)}$, so linearity of $\mathbf{M}(j\omega_0)^{-1}$ gives $V_{\mathrm{out}}^{(-)}(\omega_0) = -V_{\mathrm{out}}^{(+)}(\omega_0)$. By (H3), $\sigma_{\mathrm{pol}} = \mathrm{sign}\,\Re\{V_{\mathrm{out}}^{(+)}(\omega_0)\} \in \{+1, -1\}$ equals $\mathrm{sign}\,\lim_{\omega \to 0^+} \Re\{A_{\mathrm{dm}}(\omega)\}$, uniquely aligning low-frequency differential gain polarity without re-simulation.

\emph{(ii) Sufficient dc servo condition for non-existence of unservoed common-mode equilibrium.}
For $\mathrm{out}_1, \mathrm{out}_2 \notin \{\mathrm{VDD}, \mathrm{VSS}\}$, let $v_0 := \frac{1}{2}(x_{0,\mathrm{out}_1} + x_{0,\mathrm{out}_2})$. At dc ($s=0$), KCL across $\{\mathrm{out}_1, \mathrm{out}_2\}$ yields $I_{\mathrm{sub}}(v_0) + I_{\mathrm{servo}}(\mathbf{x}_0) = 0$ and $I_{\mathrm{servo}}(\mathbf{x}_0) = G_{\mathrm{servo}}(v_0 - V_{\mathrm{CM}})$. If $I_{\mathrm{sub}} : \R \to \R$ is $g_{\mathrm{cm}}^{\max}$-Lipschitz ($0 \le g_{\mathrm{cm}}^{\max} < G_{\mathrm{servo}}$) and $|I_{\mathrm{servo}}(\mathbf{x}_0)| > \tau_I := \frac{g_{\mathrm{cm}}^{\max} G_{\mathrm{servo}}}{G_{\mathrm{servo}} - g_{\mathrm{cm}}^{\max}}\Delta V_{\mathrm{max}}$ ($\Delta V_{\mathrm{max}} > 0$), then for every $v \in [V_{\mathrm{CM}} - \Delta V_{\mathrm{max}}, V_{\mathrm{CM}} + \Delta V_{\mathrm{max}}]$, $|v - v_0| \le \Delta V_{\mathrm{max}} + |I_{\mathrm{servo}}(\mathbf{x}_0)|/G_{\mathrm{servo}}$ yields $|I_{\mathrm{sub}}(v)| \ge (1 - g_{\mathrm{cm}}^{\max}/G_{\mathrm{servo}})|I_{\mathrm{servo}}(\mathbf{x}_0)| - g_{\mathrm{cm}}^{\max}\Delta V_{\mathrm{max}} > 0$, proving no unservoed equilibrium exists in that window. (\ctpr takes the disjunction with structural absence of passive/diode dc shunts at $\{\mathrm{out}_1, \mathrm{out}_2\}$ to cover symmetric $I_{\mathrm{servo}}(\mathbf{x}_0)=0$ stages.)
\end{proof}

\subsection{Proof of Proposition 2 (Bi-Lipschitz Margin and Submodular Hypervolume)}
\begin{proof}[Proof of Proposition~2(a)]
Fix $\lambda \ge 1$ and $0 < \epsilon_\sigma \le \sigma_j < \infty$ ($j=1,\dots,M$). Each coordinate map $\psi_{\lambda,j}(t) = (t/\sigma_j)\mathbb{I}[t \ge 0] + (\lambda t/\sigma_j)\mathbb{I}[t < 0]$ is continuous at $t=0$ and strictly increasing on $\R$ with continuous inverse $\psi_{\lambda,j}^{-1}(y) = \sigma_j y \mathbb{I}[y \ge 0] + (\sigma_j / \lambda)y \mathbb{I}[y < 0]$. For $a = b$, $\sigma_j^{-2}(a - b)^2 \le (\psi_{\lambda,j}(a) - \psi_{\lambda,j}(b))^2 \le \lambda^2 \sigma_j^{-2}(a - b)^2$ holds as $0 \le 0 \le 0$; for $a \neq b$, the symmetric chord slope $s_j(a,b) = (\psi_{\lambda,j}(a) - \psi_{\lambda,j}(b))/(a - b)$ equals $1/\sigma_j$ ($a,b \ge 0$), $\lambda/\sigma_j$ ($a,b \le 0$), or $\alpha(1/\sigma_j) + (1-\alpha)(\lambda/\sigma_j) \in [1/\sigma_j, \lambda/\sigma_j]$ with $\alpha = a/(a-b) \in (0,1)$ when $b < 0 < a$ (and symmetrically for $a < 0 < b$). Summing squared differences across $j=1,\dots,M$ yields
\[
  \Bigl(\min_j \sigma_j^{-1}\Bigr)\|\tilde{\mathbf{g}} - \tilde{\mathbf{g}}'\|_2 \;\le\; \|\psi_\lambda(\tilde{\mathbf{g}}) - \psi_\lambda(\tilde{\mathbf{g}}')\|_2 \;\le\; \Bigl(\lambda\max_j \sigma_j^{-1}\Bigr)\|\tilde{\mathbf{g}} - \tilde{\mathbf{g}}'\|_2
\]
with tight constants $L_{\min} = \min_j \sigma_j^{-1} > 0$ and $L_{\max} = \lambda\max_j \sigma_j^{-1} \le \lambda/\epsilon_\sigma < \infty$. Strict coordinate monotonicity immediately gives $\tilde{\mathbf{g}} \succ \tilde{\mathbf{g}}' \iff \psi_\lambda(\tilde{\mathbf{g}}) \succ \psi_\lambda(\tilde{\mathbf{g}}')$. At $\lambda = 0$ ($\max(\mathbf{Z},\mathbf{0})$ clipping), $\tilde{\mathbf{g}}=(1, -0.1) \succ \tilde{\mathbf{g}}'=(1, -50)$ both map to $(1/\sigma_1, 0)$, violating injectivity ($L_{\min}=0$) and strict Pareto preservation.
\end{proof}

\begin{proof}[Proof of Proposition~2(b)]
Since $1 \le |\mathcal{Y}| < \infty$, $\delta_{\mathrm{pad}} > 0$, $\epsilon_{\mathrm{top}} > 0$, and $K, Q, M \in \mathbb{Z}_{\ge 1}$, $u_{\mathrm{top},j} - r_{\mathrm{ref},j} \ge \delta_{\mathrm{pad}} + \epsilon_{\mathrm{top}} > 0$, so $w = \mathrm{Vol}([\mathbf{r}_{\mathrm{ref}}, \mathbf{u}_{\mathrm{top}}])/Q > 0$. Let $\mathcal{U} = \{1, \dots, Q\}$ and $D(\mathbf{y}) = \{r \in \mathcal{U} : \mathbf{q}_r \le \mathbf{y}\}$. Then $F_{\mathrm{HV}}(S) = w\,|\bigcup_{\mathbf{y} \in S} D(\mathbf{y})|$ satisfies $F_{\mathrm{HV}}(\emptyset) = 0$, monotonicity, and for $A \subseteq B \subseteq \mathcal{Y}$ and $\mathbf{z} \in \mathcal{Y} \setminus B$ with $U_A = \bigcup_{\mathbf{y}\in A} D(\mathbf{y}) \subseteq U_B = \bigcup_{\mathbf{y}\in B} D(\mathbf{y})$,
\[
  F_{\mathrm{HV}}(A \cup \{\mathbf{z}\}) - F_{\mathrm{HV}}(A) = w\,|D(\mathbf{z}) \setminus U_A| \;\ge\; w\,|D(\mathbf{z}) \setminus U_B| = F_{\mathrm{HV}}(B \cup \{\mathbf{z}\}) - F_{\mathrm{HV}}(B)
\]
for every $M \ge 1$. If $K \ge |\mathcal{Y}|$, greedy selection terminates in $K_{\mathrm{eff}} = |\mathcal{Y}|$ steps with $S_{\mathrm{greedy}} = \mathcal{Y}$ (exact optimum); if $1 \le K < |\mathcal{Y}|$, \citet{nemhauser1978analysis} gives $F_{\mathrm{HV}}(S_{\mathrm{greedy}}) \ge (1 - (1 - 1/K)^K)\max_{|S|\le K} F_{\mathrm{HV}}(S) \ge (1 - 1/e)\max_{|S|\le K} F_{\mathrm{HV}}(S)$. Precomputing $\mathbf{D} \in \{0,1\}^{|\mathcal{Y}|\times Q}$ in $O(|\mathcal{Y}| Q M)$ plus $K_{\mathrm{eff}} = \min(K, |\mathcal{Y}|)$ bitmask updates takes $O(|\mathcal{Y}| Q (M + \min(K, |\mathcal{Y}|))) \subseteq O(K |\mathcal{Y}| Q M)$ time.
\end{proof}

\section{10-Class Specification Contracts and Complete Ablation Matrix}
\label{sec:app_specs}

\begin{table}[h]
  \centering
  \caption{\textbf{Exact Programmatic 28nm ($V_{DD}=1.0\;\mathrm{V}$) Specification Contracts Across All 10 Functional AMS Classes in \strata} (auto-generated from \texttt{strata/sim/class\_testbenches.py}; $^*$marks primary identification gates $\mathcal{M}_{\mathrm{prim}}(c)$).}
  \label{tab:app_specs}
  \resizebox{0.99\textwidth}{!}{\begin{tabular}{lccp{8.6cm}c}
\toprule
\textbf{AMS Class $c$} & \textbf{Dim.\ $M$} & \textbf{Primary ID Gates} & \textbf{Exact 28nm ($V_{DD}=1.0\,\text{V}$) Strict Feasibility Thresholds $R_{c,j}$} & \textbf{Ver.\ Topos} \\
\midrule
\texttt{SE-OpAmp} & 8 & 4 & \texttt{Gain\_dB}$^*$ $\ge 38\,\text{dB}$, \texttt{GBW\_MHz}$^*$ $\ge 0.1\,\text{MHz}$, \texttt{PM\_deg}$^*$ $\ge 30^\circ$, \texttt{SR\_Vus} $\ge 0.2\,\text{V}/\mu\text{s}$, \texttt{CMRR\_dB} $\ge 35\,\text{dB}$, \texttt{PSRR\_dB} $\ge 32\,\text{dB}$, \texttt{Power\_uW}$^*$ $\le 1200\,\mu\text{W}$, \texttt{Area\_um2} $\le 800\,\mu\text{m}^2$ & 88 \\
\texttt{FD-OpAmp} & 8 & 4 & \texttt{Gain\_dB}$^*$ $\ge 38\,\text{dB}$, \texttt{GBW\_MHz}$^*$ $\ge 0.1\,\text{MHz}$, \texttt{PM\_deg}$^*$ $\ge 30^\circ$, \texttt{SR\_Vus} $\ge 0.2\,\text{V}/\mu\text{s}$, \texttt{CMRR\_dB} $\ge 35\,\text{dB}$, \texttt{PSRR\_dB} $\ge 32\,\text{dB}$, \texttt{Power\_uW}$^*$ $\le 1200\,\mu\text{W}$, \texttt{Area\_um2} $\le 800\,\mu\text{m}^2$ & 29 \\
\texttt{Diff-Amp} & 6 & 4 & \texttt{Gain\_dB}$^*$ $\ge 14\,\text{dB}$, \texttt{GBW\_MHz}$^*$ $\ge 1\,\text{MHz}$, \texttt{PM\_deg}$^*$ $\ge 40^\circ$, \texttt{CMRR\_dB} $\ge 24\,\text{dB}$, \texttt{Power\_uW}$^*$ $\le 800\,\mu\text{W}$, \texttt{Area\_um2} $\le 400\,\mu\text{m}^2$ & 45 \\
\texttt{SS-Amp} & 5 & 4 & \texttt{Gain\_dB}$^*$ $\ge 8\,\text{dB}$, \texttt{GBW\_MHz}$^*$ $\ge 1\,\text{MHz}$, \texttt{Swing\_V}$^*$ $\ge 0.2\,\text{V}$, \texttt{Power\_uW}$^*$ $\le 600\,\mu\text{W}$, \texttt{Area\_um2} $\le 250\,\mu\text{m}^2$ & 103 \\
\texttt{Comparator} & 6 & 3 & \texttt{Swing\_V}$^*$ $\ge 0.45\,\text{V}$, \texttt{Delay\_ns}$^*$ $\le 4.5\,\text{ns}$, \texttt{SR\_Vus} $\ge 40\,\text{V}/\mu\text{s}$, \texttt{Offset3s\_mV} $\le 25\,\text{mV}$, \texttt{Power\_uW}$^*$ $\le 500\,\mu\text{W}$, \texttt{Area\_um2} $\le 350\,\mu\text{m}^2$ & 91 \\
\texttt{LDO} & 8 & 4 & \texttt{VoutErr\_mV}$^*$ $\le 65\,\text{mV}$, \texttt{LoadReg\_mV}$^*$ $\le 65\,\text{mV}$, \texttt{LineReg\_mV} $\le 45\,\text{mV}$, \texttt{LoopGain\_dB}$^*$ $\ge 24\,\text{dB}$, \texttt{PM\_deg}$^*$ $\ge 30^\circ$, \texttt{PSRR\_dB} $\ge 24\,\text{dB}$, \texttt{Quiescent\_uW} $\le 800\,\mu\text{W}$, \texttt{Area\_um2} $\le 2000\,\mu\text{m}^2$ & 147 \\
\texttt{Bandgap} & 6 & 3 & \texttt{Vref\_V}$^*$ $\ge 0.25\,\text{V}$, \texttt{Tempco\_ppm}$^*$ $\le 320\,\text{ppm}/^\circ\text{C}$, \texttt{LineSens\_mV}$^*$ $\le 45\,\text{mV/V}$, \texttt{PSRR\_dB} $\ge 20\,\text{dB}$, \texttt{Power\_uW} $\le 500\,\mu\text{W}$, \texttt{Area\_um2} $\le 1500\,\mu\text{m}^2$ & 22 \\
\texttt{CMirror} & 5 & 3 & \texttt{MirrorErr\_pct}$^*$ $\le 12\%$, \texttt{Rout\_kOhm}$^*$ $\ge 25\,\text{k}\Omega$, \texttt{Compliance\_V} $\le 0.55\,\text{V}$, \texttt{Power\_uW}$^*$ $\le 400\,\mu\text{W}$, \texttt{Area\_um2} $\le 300\,\mu\text{m}^2$ & 110 \\
\texttt{VCO-Osc} & 5 & 2 & \texttt{OscSwing\_V}$^*$ $\ge 0.25\,\text{V}$, \texttt{Fosc\_MHz}$^*$ $\ge 5\,\text{MHz}$, \texttt{Startup\_ns} $\le 15\,\text{ns}$, \texttt{Power\_uW} $\le 2500\,\mu\text{W}$, \texttt{Area\_um2} $\le 600\,\mu\text{m}^2$ & 75 \\
\texttt{Filter-SH} & 6 & 3 & \texttt{PassGain\_dB}$^*$ $\ge -6\,\text{dB}$, \texttt{StopAtten\_dB}$^*$ $\ge 12\,\text{dB}$, \texttt{BW\_MHz}$^*$ $\ge 0.2\,\text{MHz}$, \texttt{Droop\_mV} $\le 35\,\text{mV}$, \texttt{Power\_uW} $\le 800\,\mu\text{W}$, \texttt{Area\_um2} $\le 600\,\mu\text{m}^2$ & 114 \\
\bottomrule
\end{tabular}
}
\end{table}

\begin{table}[h]
  \centering
  \caption{\textbf{Complete 11-Row Cumulative and Leave-One-Out (LOO) Ablation Matrix} ($N=110$ Balanced 11-Class Topologies, Fixed Cap $B=56$, \texttt{PTM} 28nm, \texttt{self\_supervised\_proxy}).}
  \label{tab:app_ablation}
  \small
  \begin{tabular}{lcccc}
\toprule
\textbf{Ablation Configuration ($N=110$, $B=56$)} & \textbf{Macro-F1} & \textbf{Accuracy} & \textbf{Total SPICE} & \textbf{Mean Evals / Ckt} \\
\midrule
$A_0$: Raw Unrepaired, No Orbits, No LUT & 0.1420 & 0.2000 & 5,336 & 48.51 \\
$A_1$: $+\,$\ctpr ERC \& D/S Orientation & 0.1703 & 0.2273 & 5,360 & 48.73 \\
$A_2$: $+\,$Prop.~1 Single-Probe \& CMFB (before Tier-1) & 0.1674 & 0.2273 & 5,287 & 48.06 \\
$A_3$: $+\,$Tier-1 Structural Admissibility Pruning & 0.4458 & 0.4091 & 5,476 & 49.78 \\
$A_4$: $+\,$1-WL Symmetry-Orbit Tying & 0.5486 & 0.4909 & 5,422 & 49.29 \\
\textbf{$A_5$: Full \strata ($+\,g_m/I_D$ BSIM4 LUT)} & 0.5774 & 0.5182 & 5,373 & 48.85 \\
\midrule
LOO $-\,$\ctpr ERC \& Orientation & 0.5176 & 0.4727 & 5,168 & 46.98 \\
LOO $-\,$Single-Probe (structural port order, all $B$ to sizing) & 0.5995 & 0.5455 & 5,373 & 48.85 \\
LOO $-\,$Tier-1 Structural Pruning & 0.1464 & 0.1636 & 5,273 & 47.94 \\
LOO $-\,$1-WL Symmetry Orbits & 0.5283 & 0.4909 & 5,278 & 47.98 \\
LOO $-\,g_m/I_D$ BSIM4 LUT ($\equiv A_4$, raw $(W,L)$ orbits) & 0.5486 & 0.4909 & 5,422 & 49.29 \\
\bottomrule
\end{tabular}

\end{table}

\section{Boundary Regimes, Nuanced Comparisons, and Failure Forensics}
\label{sec:app_forensics}

\begin{figure}[h]
  \centering
  \includegraphics[width=0.92\textwidth]{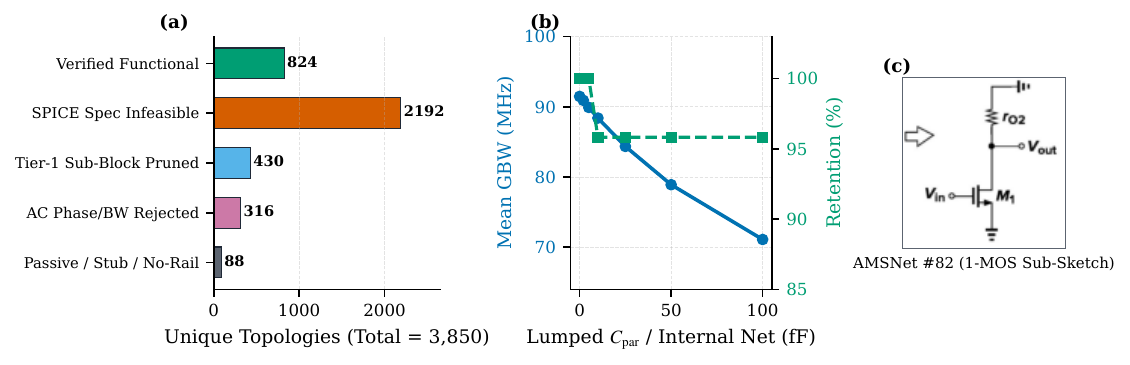}
  \caption{\textbf{Corpus Rejection Forensics and Post-Layout Parasitic Sweep.} (a)~Breakdown of all \UniqueWLTopologies{} unique \texttt{WL-1} topologies across \texttt{AMSNet~1.0} and \texttt{AnalogGenie}. (b)~\texttt{SKY130} lumped internal wire capacitance sweep $C_{\mathrm{par}} \in [0, 100]\;\mathrm{fF}$ per net across $N=24$ verified amplifiers. (c)~Real \texttt{AMSNet~1.0} \#82 single-MOSFET textbook small-signal helper sketch ($M_1 + r_{O2}$).}
  \label{fig:app_forensics}
\end{figure}

We transparently document all boundary regimes and nuanced empirical comparisons surfaced during multi-corpus evaluation:
\begin{enumerate}[leftmargin=1.5em,itemsep=3pt]
  \item \textbf{Heuristic 0-SPICE Motif Matching vs.\ Simulation-Gated Abstention (\texttt{Exp~1C}):} On pure-human \texttt{AnalogGenie} ($N=\HumanN{}$), 0-SPICE \texttt{Rule-Subgraph} achieves higher raw classification F1 (\RuleHumanFOne{}) than \strata (\StrataHumanFOne{}), because textbook chapters group partial 1--3 transistor building blocks under their parent chapter heading. \texttt{Rule-Subgraph} assigns motif labels without checking electrical operation, whereas \strata requires closed-loop BSIM4 specification satisfaction across primary physical metrics and abstains as \text{Unclassified} when a fragment cannot operate standalone. Among simulation-verified pipelines on matched 28nm CMOS, \strata improves Macro-F1 by \HumanGainOverEnhX{}$\times$--\StratSamePtmGainX{}$\times$.
  \item \textbf{Subcircuit External Pin Omission (\texttt{STRATA-Portless}):} When external \texttt{Port\#.txt} pin locations are completely hidden on \texttt{AnalogGenie} subcircuits ($\mathcal{P}_{\mathrm{ext}}=\emptyset$), Macro-F1 drops from \StrataHumanFOneFour{} to \StrataPortlessFOne{}. Unlike \texttt{AMSNet-78} schematics where external gates appear as degree-1 terminals (\CrossAmsPtmFOne{} on 28nm and \CrossAmsSkyFOne{} on \texttt{SKY130} with zero \texttt{Port\#.txt}), subcircuits connect feedback networks internally to \texttt{VOUT}, requiring boundary pin markers to distinguish internal nodes from external outputs.
  \item \textbf{Fixed-Budget Single-Probe Allocation vs.\ Structural Port Order (Table~\ref{tab:app_ablation}):} Under strict cap $B=56$, \texttt{Full STRATA} ($A_5$) spends $1$ evaluation on Proposition~1's probe at untuned $\bm{\theta}_{\mathrm{nom}}=0.45$ and $55$ on CMA-ES (`5,373` total SPICE evaluations), whereas \texttt{LOO\_Minus\_SingleProbe} spends all $56$ evaluations on CMA-ES under \ctpr's default structural port order (`5,373` evaluations), reaching \AblLooNoProbeFOne{} vs.\ \AblFullFOne{} ($60/110$ vs.\ $57/110$ accuracy, $+3$ circuits). Because open-loop AC magnitude is sign-invariant on canonical netlists, committing to a pre-sizing sign on the $\sim 5.4\%$ of differential stages where untuned biasing violates Hypothesis~H3 ($\Re\{A_{\mathrm{dm}}(\omega_0)\}\approx 0$) flips polarity on 3 boundary circuits; conversely, when pins are arbitrarily permuted or closed-loop ($N=74$ active pairs), Proposition~1 recovers \PolarityConsistPct{}\% exact sign orientation in 1 solve.
  \item \textbf{Same-Pool Interior Medoids vs.\ Frontier-Spanning Offset (\texttt{Exp~3}):} Given \strata's exact \texttt{SKY130} feasible pool ($N=18$), \phta achieves higher 15/45-corner PVT survival (\PhtaPvtFifteen{}\%/\PhtaPvtFortyFive{}\% vs.\ \SamePoolAmsqPvtFifteen{}\%/\SamePoolAmsqPvtFortyFive{}\%) at identical Pelgrom yield (\PhtaPelgromYield{}\%), whereas \texttt{AMSnet-q} Alg.~2 exhibits lower mean $V_{\mathrm{os},3\sigma}$ (\SamePoolAmsqVosMv{}\;mV vs.\ \PhtaVosMv{}\;mV) because \phta retains minimum-area Pareto corners alongside high-precision corners.
  \item \textbf{Downstream Pairwise $k$-NN Variance (\texttt{Exp~6A}) and Feasible-Pool Oracle (\texttt{Exp~6B}):} In \texttt{Exp~6A} ($N=50$), \strata-\phta warm-start (\WarmStrataEvals{} evals) significantly beats \texttt{Cold-Start Separable CMA-ES} (\WarmColdEvals{}, $p_{\mathrm{Holm}}=\WarmWilcoxonP{}$), while directional reductions over \texttt{Random-Atlas} (\WarmRandEvals{}, $p_{\mathrm{Holm}}=\WarmRandP{}$), \texttt{WL-kNN} (\WarmWlEvals{}, $p_{\mathrm{Holm}}=\WarmWlP{}$), and \texttt{AMSnet-q Medoid} (\WarmAmsqEvals{}, $p_{\mathrm{Holm}}=\WarmAmsqP{}$) are not individually significant at $N=50$. In \texttt{Exp~6B} ($N=265$), end-to-end \strata ($R^2=\EnvStrataRTwo{}$) beats single-sizing ($R^2=\EnvSingleRTwo{}$) and unfiltered \texttt{AMSnet-q} ($R^2=\EnvAmsqUnfiltRTwo{}$); providing \texttt{AMSnet-q} \pmci's pre-filtered feasible pool reaches $R^2=\EnvAmsqFeasRTwo{}$, confirming \pmci feasibility filtering as the primary driver of envelope predictability.
  \item \textbf{Corpus Rejection Forensics and Lumped Parasitic Roll-Off (\texttt{Exp~7}, Figure~\ref{fig:app_forensics}):} Across \UniqueWLTopologies{} unique topologies, $2{,}192$ fail multi-metric SPICE specs across all classes, $430$ are pruned by Tier-1 admissibility, $316$ fail un-floored AC phase/bandwidth checks, and $88$ lack rails or active devices, leaving \VerifiedAtlasTopologies{} verified topologies. Sweeping lumped internal capacitance $C_{\mathrm{par}} \in [0, 100]\;\mathrm{fF}$ across $N=24$ \texttt{SKY130} amplifiers rolls off mean GBW from \CparZeroGbw{} to \CparHundredGbw{}\;MHz while retaining \CparHundredRetain{}\% feasibility.
\end{enumerate}

\end{document}